\documentclass{article}

\usepackage{arxiv}

\usepackage[utf8]{inputenc} 
\usepackage[T1]{fontenc}    
\usepackage{url}            
\usepackage{booktabs}       
\usepackage{amsfonts}       
\usepackage{nicefrac}       
\usepackage{microtype}      
\usepackage{lipsum}
\usepackage{graphicx}
\usepackage{amssymb}
\usepackage{amsmath}
\usepackage{graphicx}
\graphicspath{{figures/}}
\usepackage{amsmath, bm}
\usepackage{amsfonts}
\usepackage{booktabs}
\usepackage{tabularx}
\usepackage[table]{xcolor}  
\usepackage{multirow}       
\usepackage{array}          
\usepackage{algorithm}
\usepackage[noend]{algpseudocode}
\usepackage{xcolor}
\usepackage{enumitem}
\usepackage{siunitx}
\usepackage{verbatim}
\usepackage{multirow} 
\usepackage{chngcntr} 
\usepackage{hyperref}
\usepackage{placeins}

\newcommand{\updatedText}[1]{\textcolor{red}{#1}}
\renewcommand{\updatedText}[1]{#1}
\newcommand{\updatedSecond}[1]{\textcolor{red}{#1}}
\renewcommand{\updatedSecond}[1]{#1}

\title{FNODE: Flow-Matching for data-driven simulation of constrained multibody systems}

\author{
 Hongyu Wang \\
  Department of Electrical and Computer Engineering\\
  University of Wisconsin-Madison\\
  Madison, WI 53706 \\
  \texttt{hwang2487@wisc.edu} \\
   \And
 Jingquan Wang \\
  Department of Mechanical Engineering\\
  University of Wisconsin-Madison\\
  Madison, WI 53706 \\
  \texttt{jwang2373@wisc.edu} \\
  \And
 Dan Negrut \\
  Department of Mechanical Engineering\\
  University of Wisconsin-Madison\\
  Madison, WI 53706 \\
  \texttt{negrut@wisc.edu} \\
}

\begin{document}
\maketitle
\begin{abstract}
	Data-driven modeling of constrained multibody dynamics remains challenged by (i) the training cost of Neural ODEs, which typically require backpropagation through an 
	ODE solver, and (ii) error accumulation in rollout predictions. We introduce a Flow-Matching Neural ODE (FNODE) framework that learns the acceleration mapping directly 
	from trajectory data by supervising accelerations rather than integrated states, turning training into a supervised regression problem and eliminating the ODE-adjoint/solver 
	backpropagation bottleneck. Acceleration targets are obtained efficiently via numerical differentiation using a hybrid fast Fourier transform (FFT) and finite-difference 
	(FD) scheme. Kinematic constraints are enforced through coordinate partitioning: FNODE learns accelerations only for the independent generalized coordinates, while the 
	dependent coordinates are recovered by solving the position-level constraint equations. We evaluate FNODE on single and triple mass--spring--damper systems, a double 
	pendulum, a slider--crank with and without friction, a vehicle model, and a cart--pole, and compare against MBD-NODE, LSTM, and fully connected baselines.
	\updatedSecond{We also compared the error contributions between numerical differentiation and Neural Network learned field mismatch in single-mass-spring-damper system, 
	which proves the error of learned acceleration field is dominated by the error of Neural Network and confirms the robustness of numerical differentiation.} 
	Across these benchmarks, FNODE achieves improved prediction accuracy and training/runtime efficiency, while maintaining constraint satisfaction through the partitioning 
	procedure. Our code and scripts are released as open source to support reproducibility and follow-on research. 
\end{abstract}






\section{Introduction}
\label{intro}


Multibody dynamics is relevant in simulation-based engineering, with applications in numerous fields, 
e.g., robotics \cite{liu2025nonlinear}, vehicle and train dynamics \cite{archut2025real,weilguny2025comparison}, biomechanics 
\cite{abdullah2024multibody}, \updatedText{aircraft dynamics \cite{zhou2025dynamic}} and molecular dynamics \cite{sun2024graph}, to name a few. Traditionally, modeling multibody systems
 has relied on first-principles approaches such as Newton-Euler formulations \cite{shabana1990dynamics}, Lagrangian mechanics \cite{bauchau2009scaling}, and
 Hamiltonian mechanics \cite{chadaj2017parallel}. While these classical methods provide accurate and interpretable models, 
 they are challenged in scenarios where the underlying dynamics are partially unknown, highly complex, or subject to unmodeled effects 
 such as friction, contact forces, or material nonlinearities \cite{flores2023contact}.

Machine learning has opened the door to data-driven simulation of dynamical systems, offering the potential to learn complex behaviors 
directly from observational data \cite{raissi2019physics,karniadakis2021physics}. Early approaches in this domain employed classical 
machine learning architectures such as LSTM \cite{vlachas2018data,hochreiter1997long} and FCNN \cite{rios1997identification} to learn discrete-time mappings from current states to future states. 
In addition, recent work \cite{go2024efficient,go2024rapidly} explored fixed-time increment DNN methods combined with dimensionality reduction techniques 
for efficient multibody dynamics simulation.
While these methods have shown promise in short-term prediction tasks, they demonstrated limited out-of-distribution generalization. 
Specifically, these black-box predictors learn fixed-timestep transitions $f: s_t \mapsto s_{t+\Delta t}$, which limits 
their temporal resolution and also causes error accumulation that grows exponentially with the prediction horizon \cite{zhu2022numerical}. 
Furthermore, such discrete-time approaches often fail to enforce fundamental conservation 
laws when applied in conjunction with conservative systems \cite{hairer2006geometric}. Physics-informed approaches have been proposed to address these limitations by 
incorporating physical constraints and laws into the model, presenting improved reliability through enforcement of physical principles 
in neural network training \cite{song2025study,song2025novel,zhang2025efficient}.
\updatedText{Neural Ordinary Differential Equations (NODEs) \cite{chen2018neural} marked a major step in continuous-time 
learning by representing system dynamics as neural vector fields integrated with standard ODE solvers. 
This formulation handles irregular time sampling and directly connects learning with established numerical 
integration theory \cite{kidger2020neural,morrill2021neural}. }

\updatedText{Meanwhile, hybrid methods combining first-principle models with NODEs have shown promise in specific applications through automatic parameter tuning \cite{zhu2024data}. However, these approaches typically assume that the functional form of the governing equations (or physics structure) is known \emph{a priori}, which can limit their ability to generalize when the underlying mechanisms are partially unknown or misspecified.}
\updatedText{To improve physical consistency, several structure-preserving learning directions have been proposed, 
including energy-based formulations such as Hamiltonian and Lagrangian Neural Networks \cite{greydanus2019hamiltonian,cranmer2020lagrangian}, 
symplectic-geometry-based approaches such as Symplectic ODE-Net and Symplectic Machine Learning (Sym-ML) frameworks 
\cite{zhong2019symplectic,song2025sym}, and methods that target numerical stability and explicit constraint satisfaction, 
e.g., Stabilized Neural Differential Equations \cite{white2023stabilized}. For flow-matching-based methods, 
recent work proposes a contrastive feature-alignment remedy for systematic failure modes that arise in the low-noise regime \cite{zeng2025flow}. 
Separately, another work develops improved training strategies for rectified flows \cite{lee2024improving}.
In a different modeling paradigm, operator-learning methods such as DeepONets \cite{lu2021deeponet} and Fourier Neural Operators \cite{li2021fno} 
learn solution operators to generalize across parameterized families of dynamical systems.}
However, despite their strengths, NODEs suffer from a bottleneck: the adjoint 
sensitivity method for backpropagation requires solving an additional backward ODE, which increases 
computational cost and memory use, and limits scalability for stiff or high-dimensional systems 
\cite{gholaminejad2019anode,wang2024mbdnode}.

Recent advances in flow-matching algorithms \cite{liu2023flow,lipman2023flow} show that vector fields 
can be learned by directly supervising derivatives at sampled points, avoiding backward ODE solves 
\cite{kim2024simulation}. The approach of learning state-acceleration mapping relationship
has been successfully applied to simulate flexible body dynamics \cite{slimak2025machine}.
Applying these ideas to constrained multibody dynamics, we train 
neural networks to approximate accelerations directly, eliminating costly integration in the 
learning loop. We propose FNODE, a framework for data-driven modeling of multibody systems. 
Instead of predicting future states via time integration, FNODE trains a neural network 
to learn directly the acceleration vector field -- the natural quantity in mechanical systems 
governed by force laws \cite{arnol1997mathematical}. Kinematic constraints are enforced through a coordinate-partitioning approach \cite{Wehage82}, in which FNODE learns the acceleration field only for the independent generalized coordinates, while the dependent coordinates are recovered implicitly by enforcing the kinematic constraint equations at the position level. This reformulation removes the need for ODE solvers in backpropagation, improving computational efficiency while preserving accuracy. By learning accelerations directly, 
the model captures the instantaneous dynamics of the system, leading to better generalization and more stable long-term predictions. Moreover, 
FNODE naturally accommodates various numerical differentiation schemes for computing 
acceleration targets from position data, including finite differences and FFT-based spectral methods, 
allowing practitioners to balance between noise robustness and accuracy based on their data characteristics \cite{li2024fourier,chen2024second}.

Directly learning accelerations requires accurate derivative information, which can be 
difficult to obtain from noisy trajectory data. To address this, we use a hybrid scheme 
that combines the accuracy of FFT-based spectral differentiation with the stability of FD near trajectory 
boundaries. Our numerical experiments show that, with proper differentiation techniques, 
the advantages of acceleration-based learning outweigh the challenges of derivative estimation.

\updatedText{The following are our main contributions in this paper:}
\begin{enumerate}
	\item We introduce FNODE, an approach for data-driven simulation of dynamical systems that learns acceleration vector fields directly from trajectory data. This removes the backpropagation bottleneck of conventional NODEs and reduces training cost while preserving accuracy.
	
	\item We present a numerical differentiation framework for constructing high-quality acceleration targets, combining FFT-based spectral differentiation with finite differences to balance accuracy and stability.
	
	\item We evaluate FNODE on a diverse set of constrained multibody dynamics benchmarks, showing consistent improvements over baselines—including  MBD-NODE, LSTMs, and FCNNs—in long-term prediction accuracy, out-of-distribution generalization, and computational efficiency.
	
	\item To support reproducibility, we provide an open-source implementation of FNODE together with benchmark problems, implementations of baseline methods, data generation tools, and evaluation protocols.
\end{enumerate}

\section{Methodology}
\label{sec:methodology}
\subsection{The Multibody Dynamics Problem}
We employ an MBD formulation in redundant generalized coordinates in which Lagrange multipliers are used to account for the presence of kinematic constraints in the equations of motion \cite{shabana2020dynamics}.
\begin{equation}
	\label{eqn: MBD}
	\begin{bmatrix}
		\mathbf{M} & \mathbf{\Phi}^T_\mathbf{q} \\
		\mathbf{\Phi_q} & 0 \\
	\end{bmatrix}
	\begin{bmatrix}
		\ddot{\mathbf{q}} \\
		\lambda \\
	\end{bmatrix}
	=
	\begin{bmatrix}
		\mathbf{F_e} \\
		\mathbf{\gamma_c}\\
	\end{bmatrix},
\end{equation}
Above, the mass matrix $\mathbf{M}$ encodes the system's inertia, while $\mathbf{\Phi_q}$ is the Jacobian of the kinematic constraints. The generalized coordinates are collected in $\mathbf{q}$, with $\ddot{\mathbf{q}}$ denoting accelerations. The Lagrange multipliers $\lambda$ correspond to the constraint forces. The right-hand side includes $\mathbf{F_e}$, the vector of external and velocity-dependent forces, and $\mathbf{\gamma_c}$, the contribution from the second time derivative of the kinematic constraints. For details, see \cite{shabana2020dynamics,Haug89}.

\subsection{Neural Ordinary Differential Equations for Multibody System Dynamics}
\subsubsection{Neural Ordinary Differential Equation (NODE)}
For a general dynamical system, the NODE framework assumes the existence of an underlying continuous-time process that governs the evolution of the system states. To that end, we assume a hidden state $\mathbf{z}(t)\in\mathbb{R}^{n_z}$ with dynamics
\updatedSecond{\begin{equation}\label{eqn: NODE}
	\frac{d\mathbf{z}(t)}{dt}=f_{\Theta}(\mathbf{z}(t),t) \; ,
\end{equation}}
where \updatedSecond{$f_{\Theta}:\mathbb{R}^{n_z}\times\mathbb{R}^{+}\to\mathbb{R}^{n_z}$} is a neural network parameterized by $\mathbf{\Theta}$. The future state $\mathbf{z}(t)$ is obtained by solving the initial value problem
\updatedSecond{\begin{equation}\label{eqn: NODE_IVP}
	\mathbf{z}(t) = \mathbf{z}(0)+\int_0^t f_{\Theta}(\mathbf{z}(\tau),\tau),d\tau = \Phi(\mathbf{z}(0),f_{\Theta},t),
\end{equation}}
with $\Phi$ denoting the chosen numerical ODE solver. The parameters $\mathbf{\Theta}$ are identified from trajectory data ${\mathbf{z}(t_i)}_{i=1}^n$. Since time steps need not be uniform, adaptive and variable-step integrators can be used, which is particularly advantageous in multibody dynamics where time scales vary widely.

\subsubsection{Flow-Matching in \updatedText{NODE}}
FNODE operates on an augmented state space defined as $\bm Z(t) = (\bm z^T(t), \dot{\bm z}^T(t))^T$ that is used to model the acceleration directly:
\updatedSecond{\begin{equation}\label{eqn: FNODE}
	\ddot {\bm z}(t,\bm{\mu}) = f_{\Theta}(\bm Z(t,\bm{\mu})) \; ,
\end{equation}}
where the initial conditions for positions and velocities are defined as:
\begin{equation}\label{eqn: i-con}
	\bm Z(0,\bm{\mu}) = (\bm z^T(0,\bm{\mu}), \dot {\bm z}^T (0,\bm{\mu}))^T \; ,
\end{equation}
and the parameter vector capturing problem-specific characteristics such as material properties, geometric parameters, or external loading conditions is
\updatedText{\[
\bm{\mu} = (\mu_1, \mu_2, \dots, \mu_{n_{\bm{\mu}}})^T \in \mathbb{R}^{n_{\bm{\mu}}} \;.
\]}
The generalized coordinates and velocity are
\updatedText{\begin{equation}\label{eqn: gen-coordinate}
	\bm z(t,\bm{\mu}) = (z^1(t,\bm{\mu}),...,z^{n_z}(t,\bm{\mu}))^T\in \mathbb{R}^{n_z},  
\end{equation}}

\updatedText{\begin{equation}
	\label{eqn: gen-velocity}
	\dot {\bm z}(t,\bm{\mu}) = (\dot {z}^1(t,\bm{\mu}),\dots,\dot {z}^{n_z}(t,\bm{\mu}))^T\in \mathbb{R}^{n_z} \; .
\end{equation}}


The neural network function processes this enriched input space that is modulated by the parameter vector $\bm{\mu}$ to predict acceleration values directly
\updatedSecond{\begin{equation}
	\label{eqn: nn}
	f_{\Theta}(\mathbf{z}(t),\bm{\mu}): \mathbb{R}^{2n_z}\times\mathbb{R}^{n_{\bm \mu}}\to \mathbb{R}^{n_z} \; .
\end{equation}}
In the inference phase, the forward integration process maintains the same mathematical structure as traditional NODEs with the definition from Eq.~\eqref{eqn: NODE_IVP} and Eq.~\eqref{eqn: FNODE}. 
\updatedSecond{\begin{equation}
	\label{eqn: FNODE-inference}
	\bm Z(t,\bm \mu)=\bm Z(0,\bm \mu)+\int_0^t f_{\Theta}(\bm Z(\tau,\bm \mu))d\tau=\Phi(\bm Z(0,\bm \mu), t)\circ f_{\Theta}=\Phi_{\Theta}(\bm Z(0,\bm \mu), t).
\end{equation}}

\begin{figure}
	\centering
	\includegraphics[width=0.75\linewidth]{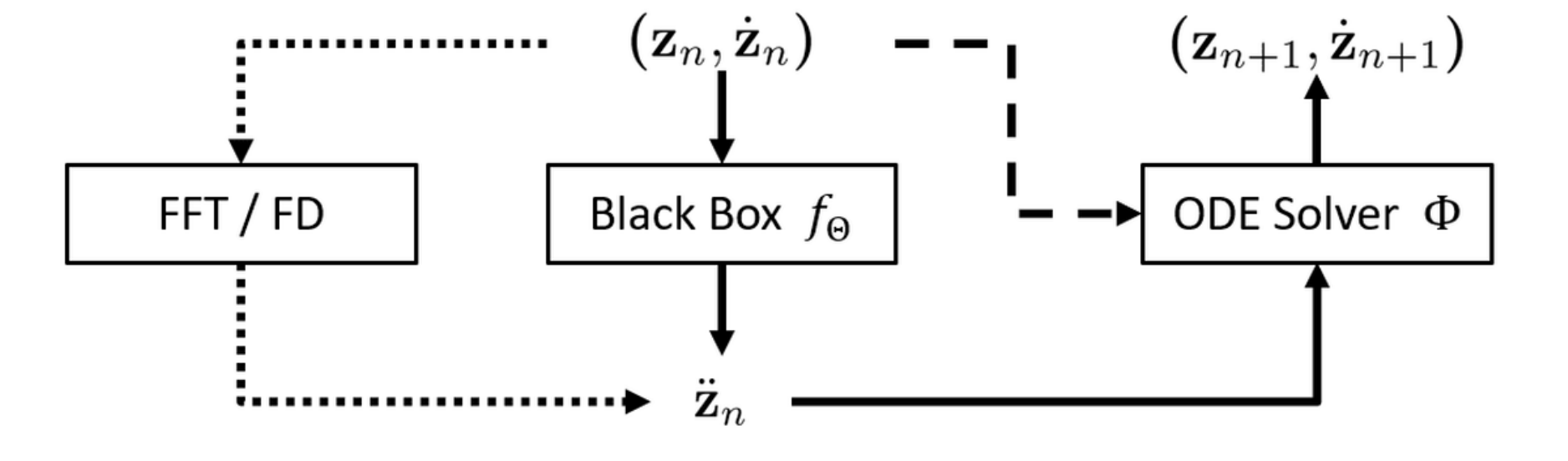}
	\caption{\updatedText{Discretized forward pass of the FNODE framework for a general multibody dynamics (MBD) system. 
	The solid arrows represent the data flow during both training and inference. 
	The dashed arrows indicate latent inference operations.
	The dotted arrows denote the calculation of time derivative (training target).
	During training, ``ground-truth'' accelerations are computed from trajectory data using FFT- or finite-difference (FD)-based differentiation. Once trained, the neural model \updatedSecond{$f_{\Theta}(\cdot)$} predicts accelerations that are supplied to an ODE solver during inference to generate the system's time evolution.}}
	\label{fig:fnode}
\end{figure}

Figure~\ref{fig:fnode} shows the discretized forward pass of FNODE. The network maps the system state to accelerations using the three-layer architecture in Table~\ref{tab:model_architecture}, with tanh activations and Xavier initialization. 

\begin{table}[htbp]
	\centering
	\caption{FNODE Architecture}
	\label{tab:model_architecture}
	\begin{tabular}{@{}lccc@{}}
		\toprule
		Layer          & Number of Neurons & Activation Function & Initialization \\
		\midrule
		Input Layer     & $2n_z$ & tanh & Xavier \\
		Hidden Layer 1  & $d_{\text{width}}$ & tanh & Xavier \\
		Hidden Layer 2  & $d_{\text{width}}$ &  tanh & Xavier \\
		Output Layer    & $n_z$ & -  & Xavier \\
		\bottomrule
	\end{tabular}
\end{table}

\subsection{Calculation of Acceleration}
\label{sec:acceleration}
Two primary approaches are employed depending on the characteristics of the available dataset: FFT spectral differentiation and FD methods.

\subsubsection{FFT Spectral Differentiation}
\label{sec:FFT-Spectral-Differentiation}
\paragraph{\updatedText{Continuous Fourier Representation}}
For dynamic systems which satisfy periodicity assumptions, FFT-based spectral differentiation provides high accuracy in \updatedText{time derivative} calculation. \updatedText{We begin by presenting the ideal mathematical framework for a smooth periodic function $z(t)$ with time interval $L = N\Delta t$ and uniform time step $\Delta t$. The} exact Fourier series representation is:
\begin{equation}
	\label{eqn: fft-series}
	\bm z(t) = \sum_{k=-\infty}^{\infty} \bm Z_k e^{2\pi i k t/L} \; .
\end{equation}
The time derivative computed by the \updatedText{exact} spectral representation assumes then the expression:
\begin{equation}
	\label{eqn: fft-gradient}
	\frac{d \bm z}{dt} = \sum_{k=-\infty}^{\infty} \frac{2\pi i k}{L} \bm Z_k e^{2\pi i k t/L} \; .
\end{equation}

\paragraph{\updatedText{Discrete Implementation and Artifacts}}
\updatedText{Eqs.~\eqref{eqn: fft-series}--\eqref{eqn: fft-gradient} describe the ideal continuous-time Fourier representation for a smooth periodic trajectory, and motivate spectral differentiation through a frequency-domain multiplier. In practice, however, we only observe a finite set of samples and compute derivatives using the discrete Fourier transform (DFT)/FFT.} 

\updatedText{In this discrete setting, sampling (finite $\Delta t$) introduces a Nyquist limit $f_N=\tfrac{1}{2\Delta t}$ and thus possible aliasing of unresolved high-frequency content, while finite-length windowing and imperfect periodicity introduce spectral leakage. These artifacts are not properties of the exact Fourier series itself, but become central once differentiation is implemented on sampled data. Moreover, since differentiation multiplies each mode by its (angular) frequency, high-frequency components---including aliased noise---are amplified in the derivative estimates. Accordingly, in the sampled DFT/FFT implementation, aliasing and finite-window effects (including Gibbs-type ringing) may introduce spurious high-frequency oscillations.} 

\updatedText{One particularly important manifestation of these discrete-sampling artifacts is the} Gibbs error\updatedText{, which} arises from \updatedText{boundary} discontinuity in \updatedText{the} periodic expansion:
\updatedText{\begin{equation}
	\label{eqn: delta}
	\Delta_{boundary}=z((N-1)\Delta t)-z(0)\neq 0 \; .
\end{equation}}
When the boundary discontinuity of magnitude $\Delta_{\text{boundary}}$ exists as defined in Eq.~\eqref{eqn: delta}, the Fourier coefficients exhibit asymptotic behavior
\begin{equation}
	\label{eqn: fft-asymptotic}
	Z(k)\sim \frac{\Delta_{boundary}}{2\pi ik/N} \; .
\end{equation}
The derivative operation multiplies each Fourier coefficient in Eq.~\eqref{eqn: fft-asymptotic} by $\frac{2\pi i k}{N\Delta t}$, resulting in
\begin{equation}
	\label{eqn: fft-asy-gradient}
	Z'(k)=\frac{2\pi ik}{N\Delta t} \cdot Z(k)\sim \frac{\Delta_{boundary}}{2\Delta t}\; .
\end{equation}
This implies that, in the presence of a boundary discontinuity, the differentiated Fourier modes do not decay with $|k|$ in the high-frequency range; instead, their magnitude remains $O\!\left(\tfrac{|\Delta_{\text{boundary}}|}{\Delta t}\right)$ as $|k|\to\infty$ (cf. Eq.~\eqref{eqn: fft-asy-gradient}). As a result, the truncated Fourier reconstruction of the derivative can exhibit spurious high-frequency oscillations (Gibbs-type ringing), with a characteristic (order-of-magnitude) error scale
\begin{equation}
	\label{eqn: gibbs}
	\epsilon_{\text{Gibbs}} \sim \frac{|\Delta_{\text{boundary}}|}{2\Delta t} \; .
\end{equation}

The error is independent of $N$ but inversely proportional to the time step, which means it would be particularly problematic for fine temporal resolution. 

To \updatedText{address} the Gibbs phenomenon problem described in Eq.~\eqref{eqn: gibbs}\updatedText{ and other boundary discontinuities}, \textit{least-square detrending} is applied\updatedText{:}
\begin{equation}
	\label{eqn: lsd-l}
	\ell(t) \;=\; a + bt, 
	\quad
	\begin{bmatrix}
		a & b
	\end{bmatrix}
	\;=\;
	\operatorname*{argmin}_{a,b}\;
	\sum_{i}\!\bigl(z_i - (a + b\,t_i)\bigr)^2 \; ,
\end{equation}
where $\ell(t)$ is the linear trend function to approximate the trend of the raw dataset. The detrended dataset is formulated using Eq.~\ref{eqn: lsd-l} as
\begin{equation}
	\label{eqn: lsd-z}
	z_d(t) \;=\; z(t) - \ell(t) \; .
\end{equation}
The new discontinuity value is formulated from Eq.~\eqref{eqn: lsd-z} as
\begin{equation}
	\label{eqn: delta-new}
	\Delta_{new}
	\;=\;
	z_d(N-1) - z_d(0)
	\;=\;
	\Delta_{\text{boundary}} - b\, L \; .
\end{equation}
Even though the detrend narrows the gap of the two ends of dataset as described in Eq.~\eqref{eqn: delta-new}, the \updatedText{first-order derivative} of the dataset is still not continuous. In order to further reduce Gibbs error while maintain the spectrum information as much as possible, we adopt \textit{mirror reflection} and \textit{cosine taper}.
\begin{equation}
	\label{eqn: mirror_all}
	\begin{aligned}
		z_e(t) &=
		\left\{
		\begin{array}{ll}
			\tau_L(t)\,z_d(-t),       & -M \le t < 0, \\[4pt]
			z_d(t),                   & 0 \le t \le N, \\[4pt]
			\tau_R(t)\,z_d(2N - t),   & N < t \le N+M,
		\end{array}
		\right.
		\hspace{2em}
		M = \dfrac{N}{4}, \\[6pt]
		\tau(t) &= \tfrac{1}{2} \left(1 - \cos\left(\dfrac{\pi t}{M}\right)\right),
		\quad t = 0, 1, \dots, M.
	\end{aligned}
\end{equation}
Above, $\tau(i)$ is the cosine window weight and $M$ is the length of Mirror Reflection. The amplitude of the dataset can be smoothly transited through this process. A Tukey window is then applied, tapering the signal to zero at both ends so that it appears periodic and suitable for FFT-based differentiation:
\begin{equation}
	\label{eqn: window}
	w(t) \;=\;
	\begin{cases}
		\dfrac12\!\Bigl[\,1+\cos\!\Bigl(\pi\,
		\dfrac{\alpha L/2 - t}{\alpha L/2}\Bigr)\Bigr],
		& 0 \le t < \tfrac{\alpha L}{2},\\[8pt]
		1,
		& \tfrac{\alpha L}{2} \le t \le L - \tfrac{\alpha L}{2},\\[8pt]
		\dfrac12\!\Bigl[\,1+\cos\!\Bigl(\pi\,
		\dfrac{t-(L-\alpha L/2)}{\alpha L/2}\Bigr)\Bigr],
		& L-\tfrac{\alpha L}{2} < t \le L,
	\end{cases}
	\quad
	\alpha = 0.2
\end{equation}

Detrending, mirror reflection, and Tukey windowing smooth structural discontinuities, but FFT differentiation still amplifies residual noise and high-frequency artifacts. To suppress these effects, we apply a Gaussian low-pass filter in the frequency domain, defined as
\begin{equation}
	\label{eqn: gaussian}
	G(k)
	\;=\;
	\exp\!\Bigl[-\tfrac12\!\Bigl(\tfrac{k}{\sigma}\Bigr)^{2}\Bigr],
	\quad
	\sigma \;\approx\; \dfrac{N}{20} \; ,
\end{equation}
where $k$ is the frequency index and $\sigma$ controls the filter bandwidth. The choice of $\sigma \approx N/20$ ensures that the filter effectively suppresses high-frequency noise while preserving the essential spectral content of the signal.

\paragraph{\updatedText{Complete FFT-Based Differentiation Pipeline}}
The complete FFT-based differentiation process is formulated as follows. 
\updatedText{Let $t_n=n\Delta t$ for $n=0,1,\dots,N-1$, and denote the sampled (preprocessed) sequence by $\mathbf{z}_e[n]:=\mathbf{z}_e(t_n)$.}
First, we compute the discrete Fourier transform (DFT) of the preprocessed signal $\mathbf{z}_e[n]$ with the applied window function $w[n]$ from Eq.~\eqref{eqn: window}:
\begin{subequations}
	\begin{equation}
		\label{eqn: fft-detrended}
		\mathbf{Z}_e(k) = \sum_{n=0}^{N-1} \mathbf{z}_e[n] \, w[n] \, e^{-j2\pi kn/N} \; .
	\end{equation}
	Next, we apply the differentiation operator in the frequency domain, together with the Gaussian filter $G(k)$ from Eq.~\eqref{eqn: gaussian}:
	\begin{equation}
		\label{eqn: fft-derivative}
		\mathbf{Z}_e'(k) = j\omega_k \, \mathbf{Z}_e(k) \, G(k) \; , 
	\end{equation}
	where $\omega_k = \tfrac{2\pi k}{N\Delta t}$ is the angular frequency. 
	Finally, the filtered time-domain derivative is obtained by the inverse FFT:
	\begin{equation}
		\label{eqn: ifft}
		\dot{\mathbf{z}}[n] = \mathcal{F}^{-1}\!\big[\mathbf{Z}_e'(k)\big] 
		= \frac{1}{N}\sum_{k=-N/2}^{N/2-1} \mathbf{Z}'_e(k)\, e^{j2\pi kn/N} \; .
	\end{equation}
\end{subequations}
This approach combines spectral differentiation with noise suppression.
\updatedText{The FFT-based pipeline is most reliable for smooth, near-periodic trajectories 
after preprocessing; for non-periodic or non-smooth signals (e.g., impacts/contact events), Fourier truncation can lead to 
Gibbs oscillations and boundary artifacts, and any explicit differentiation operator (FFT or FD) amplifies measurement noise. 
Therefore, when FNODE is trained using explicit acceleration labels obtained via differentiation, the current differentiation choices 
primarily limit applicability to smooth regimes. Note that practical error diagnostics for differentiated signals, including a reconstruction-consistency check based on integrating the derivatives and comparing against the original trajectory, are described in Section~\ref{sec:FD-method}.} 

\updatedText{Since acceleration labels are obtained via explicit differentiation, occasional high-frequency artifacts
(e.g., due to boundary treatment, FFT/finite-difference effects, or nonsmooth events such as frictional
transitions) can manifest as outliers in the target residual. In such cases, replacing the mean-squared
error (MSE) with a robust regression loss can reduce sensitivity to outliers without changing the FNODE
architecture. A common choice is the Huber loss applied to the component-wise residual $r = \hat{a} - a$:}
\updatedText{\begin{equation}
\label{eq:huber_loss}
\rho_\delta(r)=
\begin{cases}
\tfrac12 r^2, & |r|\le \delta,\\[4pt]
\delta\left(|r|-\tfrac12\delta\right), & |r|>\delta.
\end{cases}
\end{equation}
}
\updatedText{For small residuals the loss behaves like MSE, while large residuals are down-weighted (linear growth),
making training less sensitive to impulsive artifacts. In this manuscript we focus on noise-free
trajectories for clarity and do not perform a systematic study over injected noise types/levels (e.g.,
Gaussian or impulse noise); however, robust losses such as \eqref{eq:huber_loss} are a standard mitigation
when differentiated targets contain occasional spikes.}



\subsubsection{Finite Difference Method}
	\label{sec:FD-method}
	\updatedText{For non-periodic datasets, finite-difference (FD) schemes provide a simple and robust alternative to FFT-based differentiation. Let \updatedSecond{$\dot{z}_t$} denote the FD estimate of \updatedSecond{$\dot{z}(t)$}. Using the central-difference formula, \updatedSecond{$\dot{z}_t=\tfrac{z(t+\Delta t)-z(t-\Delta t)}{2\Delta t}$}, and thus \updatedSecond{$\dot{z}(t)=\dot{z}_t+\mathcal{O}(\Delta t^2)$} \cite{hildebrand1987introduction}. At the boundaries, we use one-sided formulas, \updatedSecond{$\dot{z}_t=\tfrac{z(t+\Delta t)-z(t)}{\Delta t}$} and \updatedSecond{$\dot{z}_t=\tfrac{z(t)-z(t-\Delta t)}{\Delta t}$}, and hence \updatedSecond{$\dot{z}(t)=\dot{z}_t+\mathcal{O}(\Delta t)$} in either case. Although less accurate than spectral differentiation, FD methods are more stable for short trajectories and at dataset boundaries, making them complementary in our hybrid FFT--FD approach.}
		
	\updatedText{Higher-order FD discretizations are also available. For example, let \updatedSecond{$\dot{z}^{(5)}_t$} denote the five-point central FD estimate, 
	\updatedSecond{\[
	\dot{z}^{(5)}_t=\tfrac{-z(t+2\Delta t)+8z(t+\Delta t)-8z(t-\Delta t)+z(t-2\Delta t)}{12\Delta t} \; ,
	\]}
	so that \updatedSecond{$\dot{z}(t)=\dot{z}_t^{(5)}+\mathcal{O}(\Delta t^4)$}, together with matching higher-order one-sided stencils near boundaries. In our setting, trajectories are short and derivative labels are obtained from sampled data; wider stencils increase the number of samples affected by boundary treatment and can amplify high-frequency noise. We therefore adopt the three-point central stencil in the interior and one-sided stencils at the endpoints as a conservative accuracy--robustness tradeoff.}
	
	\updatedText{To obtain a practical estimate of differentiation error on a given dataset, one may use two complementary consistency checks. First, compute derivative estimates with two stencil orders and define a relative disagreement indicator \updatedSecond{$\eta'=\tfrac{\|\dot{z}_t^{(5)}-\dot{z}_t^{(3)}\|_2}{\|\dot{z}_t^{(5)}\|_2}$}, where \updatedSecond{$\dot{z}_t^{(3)}$} and \updatedSecond{$\dot{z}_t^{(5)}$} denote the three-point and five-point FD estimates of \updatedSecond{$\dot{z}(t)$} (and analogously for \updatedSecond{$\ddot{z}(t)$}), and $\|\cdot\|_2$ denotes the discrete $\ell_2$ norm over the sampled trajectory. Second, define a reconstruction-consistency indicator by integrating the differentiated signals and comparing to the original trajectory, e.g., \updatedSecond{$\tilde z_{t+\Delta t}=\tilde z_t+\Delta t\,\dot{z}_t$} with $\tilde z_0=z_0$, and form the normalized reconstruction residual \updatedSecond{$\eta_z=\tfrac{\|z_t-\tilde z_t\|_2}{\|z_t\|_2}$}. These diagnostics provide a practical way to quantify sensitivity to the differentiation scheme and to detect differentiation artifacts when constructing acceleration targets for FNODE training; a systematic evaluation of $\eta'$ and $\eta_z$ across datasets is deferred to future work and falls outside the scope of the present contribution.}

\updatedText{\subsection{Qualitative Analysis of Error Propagation in Integration}}\label{sec:error-propagation}
While FNODE avoids numerical integration during training, its long-term accuracy depends on how errors propagate through the ODE solver. To carry out a basic error analysis following a classical framework \cite{hairer1993classical,brenan1996numerical}, let the true system state at step $n$ be $Z_n$, advanced by the exact evolution operator \updatedText{$\Phi$}, and let $\hat{Z}_n$ denote the predicted state advanced by the learned solver \updatedText{$\Phi_\Theta$}. The global error is $E_n = \hat{Z}_n - Z_n$. The exact and predicted updates are
	\updatedText{\[
	Z_{n+1} = \Phi(Z_n), \qquad \hat{Z}_{n+1} = \Phi_\Theta(\hat{Z}_n),
	\]}
so that
	\updatedText{\[
	E_{n+1} = \hat{Z}_{n+1} - Z_{n+1} = \Phi_\Theta(\hat{Z}_n) - \Phi(Z_n).
	\]}
Substituting $\hat{Z}_n = Z_n + E_n$ gives
	\updatedText{\[
	E_{n+1} = \Phi_\Theta(Z_n + E_n) - \Phi(Z_n).
	\]}
	A first-order Taylor expansion around $Z_n$ yields
	\updatedText{\[
	\Phi_\Theta(Z_n + E_n) = \Phi_\Theta(Z_n) +
	\frac{\partial \Phi_\Theta}{\partial Z}\bigg|_{Z_n} E_n + \mathcal{O}(\lVert E_n \rVert^2)\; ,
	\]}
where \updatedText{$\tfrac{\partial \Phi_\Theta}{\partial Z}\big|_{Z_n}$} denotes the Jacobian of the learned flow map \updatedText{$\Phi_\Theta$} with respect to the state, evaluated at $Z_n$. Consequently, one obtains
	\updatedText{\[
	E_{n+1} = \underbrace{\big(\Phi_\Theta(Z_n) - \Phi(Z_n)\big)}_{\text{local error}}
	+ \underbrace{\frac{\partial \Phi_\Theta}{\partial Z}\bigg|_{Z_n} E_n}_{\text{propagated error}}
	+ \mathcal{O}(\lVert E_n \rVert^2)\; .
	\]}
\updatedText{Qualitatively, each step combines a local error (discrepancy between $\Phi_\Theta$ and $\Phi$) and a propagated error (amplification of the previous error by the solver Jacobian), showing how small inaccuracies in the learned acceleration field can accumulate over time, and thus underscoring the need for accurate acceleration targets during training.}
	
\updatedText{To connect integration error with training error and differentiation-induced label error, 
we decompose the one-step defect. Let the true augmented dynamics be \updatedSecond{$\ddot z = f(Z)$} and let \updatedSecond{$\hat{f}_\Theta(Z)$} 
denote the learned vector field (\updatedSecond{FNODE trained on numerically differentiated acceleration labels}). Let \updatedSecond{$\Psi_{\Delta t}(\cdot;f)$} denote one step of a numerical 
integrator with step size $\Delta t$ applied to the vector field \updatedSecond{$f$}. In this view, 
\updatedSecond{$\Phi(Z)=\Psi_{\Delta t}(Z;f)$} and \updatedSecond{$\Phi_\Theta(Z)=\Psi_{\Delta t}(Z;\hat{f}_\Theta)$} 
represent the corresponding one-step maps. The true and learned rollouts satisfy 
\updatedSecond{$Z_{n+1}=\Psi_{\Delta t}(Z_n;f)$} and \updatedSecond{$\widehat Z_{n+1}=\Psi_{\Delta t}(\widehat Z_n;\hat f_\Theta)$}, 
and the global error $E_n=\widehat Z_n-Z_n$ evolves as \updatedSecond{$E_{n+1}=\Psi_{\Delta t}(Z_n+E_n;\hat f_\Theta)-\Psi_{\Delta t}(Z_n;f)$}}.
	
\updatedText{Adding and subtracting \updatedSecond{$\Psi_{\Delta t}(Z_n;\hat f_\Theta)$ }yields
	\updatedSecond{$E_{n+1}=\delta_n+\bigl(\Psi_{\Delta t}(Z_n+E_n;\hat f_\Theta)-\Psi_{\Delta t}(Z_n;\hat f_\Theta)\bigr)$},
where \updatedSecond{$\delta_n:=\Psi_{\Delta t}(Z_n;\hat f_\Theta)-\Psi_{\Delta t}(Z_n;f)$} is the one-step defect at the true state. \updatedSecond{Under standard smoothness assumptions on the true dynamics and assuming the use of a consistent order-$p$ 
numerical integrator whose induced numerical flow satisfies a Lipschitz-type stability bound over the rollout region, the one-step defect can be bounded as:}
	\updatedSecond{$\|\delta_n\|\le C_{\mathrm{int}}\Delta t^{p+1}+C_{\mathrm{fld}}\Delta t\,\|\hat f_\Theta(Z_n)-f(Z_n)\|$}, i.e., it combines the local truncation error of the time integrator and the effect of vector-field mismatch over one step.}
	
\updatedText{The learned-field error has two contributions. FNODE is trained on numerically differentiated acceleration labels 
\updatedSecond{$\tilde{\ddot{z}}=\ddot{z}+e_{\mathrm{diff}}$}, so \updatedSecond{$\hat f_\Theta-f=(\hat f_\Theta-\tilde f)+(\tilde f-f)$}, where \updatedSecond{$\tilde f$} is the 
vector field consistent with the differentiated labels. The first term is controlled by the training objective; in particular, 
if \updatedSecond{$L(\Theta)=\tfrac1N\sum_i\|\hat f_\Theta(Z_i)-\tilde f(Z_i)\|_2^2$}, then $\varepsilon_{\mathrm{train}}:=\sqrt{L(\Theta)}$ 
provides an RMS proxy for the learned-field mismatch on the training distribution. The second term reflects differentiation 
artifacts and can be assessed through the diagnostics introduced in Section~\ref{sec:FD-method}, such as the stencil-disagreement 
indicator $\eta'$ and the trajectory reconstruction residual $\eta_z$.}
	
\updatedText{Finally, the second term in $E_{n+1}$ captures error propagation through integration. If the integrator is stable and satisfies
	\updatedSecond{$\bigl\|\tfrac{\partial}{\partial Z}\Psi_{\Delta t}(Z;\hat f_\Theta)\bigr\|\le 1+C\Delta t$}
	over the rollout region, then
	\updatedSecond{$\|\Psi_{\Delta t}(Z_n+E_n;\hat f_\Theta)-\Psi_{\Delta t}(Z_n;\hat f_\Theta)\|\le (1+C\Delta t)\|E_n\|$},
which implies the recursion $\|E_{n+1}\|\le (1+C\Delta t)\|E_n\|+\|\delta_n\|$. Consequently, $\|E_{n}\|$ grows at most like $(1+C\Delta t)^n$, and for $T=n\Delta t$ one obtains
	$\|E(T)\|\lesssim \exp(CT)\bigl(\|E(0)\|+\sum_k\|\delta_k\|\bigr)$.
This shows explicitly how integration error accumulation couples to (i) the integrator order and step size through $C_{\mathrm{int}}\Delta t^{p+1}$, (ii) 
the training error through $\varepsilon_{\mathrm{train}}$, and (iii) differentiation-induced label error through the diagnostics $\eta'$ and $\eta_z$.}
\updatedSecond{For the three error sources, (i) could be illustrated by the trajectory error, while (ii) and (iii) represent the vector-field mismatch in acceleration.
These three error sources are further analyzed below.
}

\updatedSecond{\subsubsection{Integrator error}}
\updatedSecond{For a $p$th-order accurate time integrator, the global trajectory-level MSE is 
represented as:
\[
\widetilde{\mathrm{MSE}}_Z(\Delta t)
=\frac{1}{N}\sum_{n=0}^{N-1}\big\|\Phi(Z_n)-Z_{n+1}\big\|^2,
\]
where $\Phi(Z_n)$ is the one-step update produced using the ground truth acceleration with step size $\Delta t$.
As mentioned above, the single-step truncation error is $\mathcal{O}(\Delta t^{p+1})$, 
so the error over $t\in[0,T]$ is $\mathcal{O}(\Delta t^{p})$ uniformly, then
\[
\mathrm{MSE}_Z(\Delta t)\sim\mathcal{O}(\Delta t^{2p}),
\]
Note that since RK4 has $p = 4$, the global discretization error (in the state) over a fixed time horizon typically scales as $\mathcal{O}(\Delta t^{4})$ under standard smoothness assumptions.}

\updatedSecond{\subsubsection{Neural Network learned field mismatch}}
\updatedSecond{We train FNODE with analytical accelerations $\ddot{z}$ and acceleration calculated by FD $\tilde{\ddot{z}}$ as 
targets and evaluate the predicted accelerations $\mathring{\ddot{z}}_n$ (trained on $\ddot{z}$) 
and $\hat{\ddot{z}}$ (trained on $\tilde{\ddot{z}}$) on the ground truth $\ddot{z}_n$, reporting
$\mathring{\mathrm{MSE}}_{\ddot{z}}(\Delta t)$ and $\widehat{\mathrm{MSE}}_{\ddot{z}}(\Delta t)$:
\[
    \mathring{\mathrm{MSE}}_{\ddot{z}}(\Delta t) = \frac{1}{N}\sum_{n=0}^{N-1}\|\mathring{\ddot{z}}_n-\ddot{z}_n\|^2
\]
\[
    \widehat{\mathrm{MSE}}_{\ddot{z}}(\Delta t) = \frac{1}{N}\sum_{n=0}^{N-1}\|\hat{\ddot{z}}_n-\ddot{z}_n\|^2
\]}

\updatedSecond{\subsubsection{Differentiation-induced label error}}
\updatedSecond{To interpret FD slopes, we partition indices into an interior set $\mathcal{I}$ and a boundary
set $\mathcal{B}=\{0,N-1\}$, where $|\mathcal{B}|=2$ is a constant independent of $N$ (the samples where
one-sided stencils are used) and $\mathcal{I}=\{1,\dots,N-2\}$, where $|\mathcal{I}|=N-2$.
Define the pointwise FD acceleration error $\tilde{e}_n=\tilde{\ddot{z}}_n-\ddot{z}_n$.
Let $q_{\mathcal{I}}$ and $q_{\mathcal{B}}$ denote the order of the finite-difference method used to compute accelerations
in the interior and boundary regions, respectively, i.e.,
$\|\tilde{e}_n\|=\mathcal{O}(\Delta t^{q_{\mathcal{I}}})$ for $n\in\mathcal{I}$ and
$\|\tilde{e}_n\|=\mathcal{O}(\Delta t^{q_{\mathcal{B}}})$ for $n\in\mathcal{B}$.
Here $\mathcal{I}$ contains $N-\mathcal{O}(1)$ interior time indices, whereas $\mathcal{B}$ contains only $\mathcal{O}(1)$
boundary indices (e.g., the first/last few samples required by the stencil). Over a fixed time horizon $T$, the number of
time samples satisfies $N \sim T/\Delta t$.}

\updatedSecond{Then the interior and boundary contributions to acceleration MSE scale as
\begin{equation}
\label{eqn:MSE_I}
\begin{aligned}
\widetilde{\mathrm{MSE}}_{\ddot{z},\mathcal{I}}(\Delta t)
&=\frac{1}{N}\sum_{n\in\mathcal{I}}\|\tilde{e}_n\|^2\\
&\sim\frac{|\mathcal{I}|}{N}\,\mathcal{O}_{\mathcal{I}}(\Delta t^{2q_{\mathcal{I}}})
=\frac{N-2}{N}\mathcal{O}_{\mathcal{I}}(\Delta t^{2q_{\mathcal{I}}})
=(1-\frac{2}{N})\mathcal{O}_{\mathcal{I}}(\Delta t^{2q_{\mathcal{I}}}),
\end{aligned}
\end{equation}}

\updatedSecond{\begin{equation}
\label{eqn:MSE_B}
\begin{aligned}
\widetilde{\mathrm{MSE}}_{\ddot{z},\mathcal{B}}(\Delta t)
=\frac{1}{N}\sum_{n\in\mathcal{B}}\|\tilde{e}_n\|^2
\sim\frac{|\mathcal{B}|}{N}\,\mathcal{O}_{\mathcal{B}}(\Delta t^{2q_{\mathcal{B}}})
=\frac{2}{N}\mathcal{O}_{\mathcal{B}}(\Delta t^{2q_{\mathcal{B}}})
\end{aligned}
\end{equation}}

\updatedSecond{Since $N-1=\frac{T}{\Delta t}$ and the physical time is not extremely long ($10^6$ or above), 
$\frac{1}{N-1}=\frac{\Delta t}{T}\sim \mathcal{O}(\Delta t)$. Besides, as $N\gg1$, 
$$\frac{2}{N}\approx\frac{2}{N-1}=2\frac{\Delta t}{T}\sim \mathcal{O}(\Delta t)$$}

\updatedSecond{Then we can rewrite Eq.(\ref{eqn:MSE_B}) as:
\begin{equation}
\label{eqn:MSE_B_2}
 \widetilde{\mathrm{MSE}}_{\ddot{z},\mathcal{B}}(\Delta t)
=\frac{1}{N}\sum_{n\in\mathcal{B}}\|\tilde{e}_n\|^2
\sim\frac{|\mathcal{B}|}{N}\,\mathcal{O}_{\mathcal{B}}(\Delta t^{2q_{\mathcal{B}}})
=\frac{2}{N}\mathcal{O}_{\mathcal{B}}(\Delta t^{2q_{\mathcal{B}}})
\approx \mathcal{O}_{\mathcal{B}}(\Delta t^{2q_{\mathcal{B}}+1})
\end{equation}}

\updatedSecond{The global FD acceleration MSE is:}

\updatedSecond{\begin{equation}
\label{eqn:accel_mse_overall}
\widetilde{\mathrm{MSE}}_{\ddot{z}}(\Delta t)
=\widetilde{\mathrm{MSE}}_{\ddot{z},\mathcal{I}}(\Delta t)
+\widetilde{\mathrm{MSE}}_{\ddot{z},\mathcal{B}}(\Delta t),
\end{equation}}

\updatedSecond{The overall MSE in Eq.~(\ref{eqn:accel_mse_overall}) is dominated, for sufficiently small $\Delta t$,
by the term with the smaller exponent in $\Delta t$ (i.e., the term that decays more slowly and therefore yields a larger error).
Therefore, according to Eq.~(\ref{eqn:MSE_I}) and Eq.~(\ref{eqn:MSE_B_2}), if $2q_{\mathcal{I}}<2q_{\mathcal{B}}+1$,
the interior error dominates, and vice versa.}

\updatedSecond{To simplify step-size convergence order comparisons, the root mean squared error (RMSE) can also be used to quantify the error.
Although RMSE makes it difficult to directly separate the contributions from $\mathcal{I}$ and $\mathcal{B}$, its effect on the observed
convergence order is universal. Therefore, the order of $\widetilde{\mathrm{RMSE}}_{\ddot{z},\mathcal{I}}(\Delta t)$ is $q_{\mathcal{I}}$,
whereas the order of $\widetilde{\mathrm{RMSE}}_{\ddot{z},\mathcal{B}}(\Delta t)$ is $q_{\mathcal{B}}+\tfrac{1}{2}$.}

\updatedSecond{For a fixed-order time integrator with a fixed step size, any acceleration error, regardless of its source, enters the same numerical update operator at each step. Therefore, when comparing rollouts generated with the same integrator and $\Delta t$, differences observed at the trajectory level primarily reflect differences in the underlying acceleration (vector-field) errors, rather than differences in the integrator. Consequently, to assess the robustness of our numerical differentiation, it is necessary to compare, under identical integration settings, the learned vector-field mismatch against the numerical-differentiation error.}

\updatedSecond{The quantitative error-propagation analysis that isolates the contribution of each source is provided in Section~\ref{sec:single_mass_spring_damper}.}

\subsection{Loss Function and Optimization}
\label{sec:loss}
The FNODE loss directly supervises accelerations. Given predicted accelerations $\hat{\ddot{\mathbf{z}}}_i = f(\mathbf{z}_i,\dot{\mathbf{z}}_i;\mathbf{\Theta})$ and reference accelerations $\ddot{\mathbf{z}}_i$ obtained from trajectory data via FFT- or FD-based differentiation, we minimize the mean squared error (MSE)
\updatedText{\begin{equation}\label{eqn:loss_traj}
	\mathcal{L}(\mathbf{\Theta}) = \frac{1}{N}\sum_{i=0}^{N-1} \lVert \hat{\ddot{\mathbf{z}}}_i - \ddot{\mathbf{z}}_i \rVert^2_2 = \frac{1}{N}\sum_{i=0}^{N-1} \lVert f(\mathbf{z}_i,\dot{\mathbf{z}}_i;\mathbf{\Theta}) - \ddot{\mathbf{z}}_i \rVert^2_2.
\end{equation}}
The optimal parameters are obtained as
\begin{equation}\label{eqn: opt_unconstraints}
	\mathbf{\Theta}^* = \arg\min_{\mathbf{\Theta}} \mathcal{L}(\mathbf{\Theta}).
\end{equation}
We train FNODE using the Adam optimizer \cite{Kingma2014AdamAM}. The training algorithm is shown in \ref{sec:alg}. A discussion of the training costs is deferred to Section~\ref{sec:discussion}.

\section{Numerical Experiments}
\label{sec:numericalexperiments}








\updatedText{\subsection{Single-Mass-Spring System}\label{sec:single_mass_spring}}

\updatedText{
An undamped single mass--spring system serves as an energy-conserving benchmark for long-horizon prediction. Its dynamics are governed by
	\begin{equation}\label{eqn:sms}
		\ddot{x}=-\tfrac{k}{m}x \;,
	\end{equation}
where $x$ denotes the displacement from equilibrium, $m=10$ kg, and $k=50$ N/m. The system is initialized with $x(0)=1$ m and $\dot{x}(0)=0$, and simulated with step size $\Delta t=0.01$ s for 3000 steps (30 s). The first 300 steps are used for training and the remaining 2700 steps are reserved for extrapolation.
}

\updatedText{
To highlight the system's energy-conserving structure, we also express the dynamics in Hamiltonian form. Let $q=x$ be the generalized position and $p=m\dot{q}$ be the generalized momentum. The total energy (Hamiltonian) is
}
\updatedText{
\begin{equation}\label{eqn:sms_H}
	T(p)=\frac{p^2}{2m},\qquad V(q)=\tfrac{1}{2}kq^2,\qquad H(p,q)=T(p)+V(q) \;,
\end{equation}
}
\updatedText{which induces Hamilton's equations}
\updatedText{
\begin{equation}\label{eqn:sms_hamilton}
	\dot{q}=\frac{\partial H}{\partial p}=\frac{p}{m},\qquad
	\dot{p}=-\frac{\partial H}{\partial q}=-kq  \;.
\end{equation}
}

\updatedText{
We compare FNODE against structure-preserving baselines, including the Hamiltonian Neural Network (HNN) and Lagrangian Neural Network (LNN) \cite{wang2024mbdnode}. HNN learns a scalar Hamiltonian $H_\theta(p,q)$ and obtains time derivatives through Hamilton's equations, while LNN learns a Lagrangian $L_\theta(q,\dot{q})=T(\dot{q})-V(q)$ 
and predicts accelerations via the Euler--Lagrange equation. We also consider the Runge-Kutta 4th order (RK4) integrator as a baseline, 
which infers the future steps through the exact dynamics from Eq.~\eqref{eqn:sms} without learning.
}

\begin{figure}
	\centering
	\includegraphics[width=12cm]{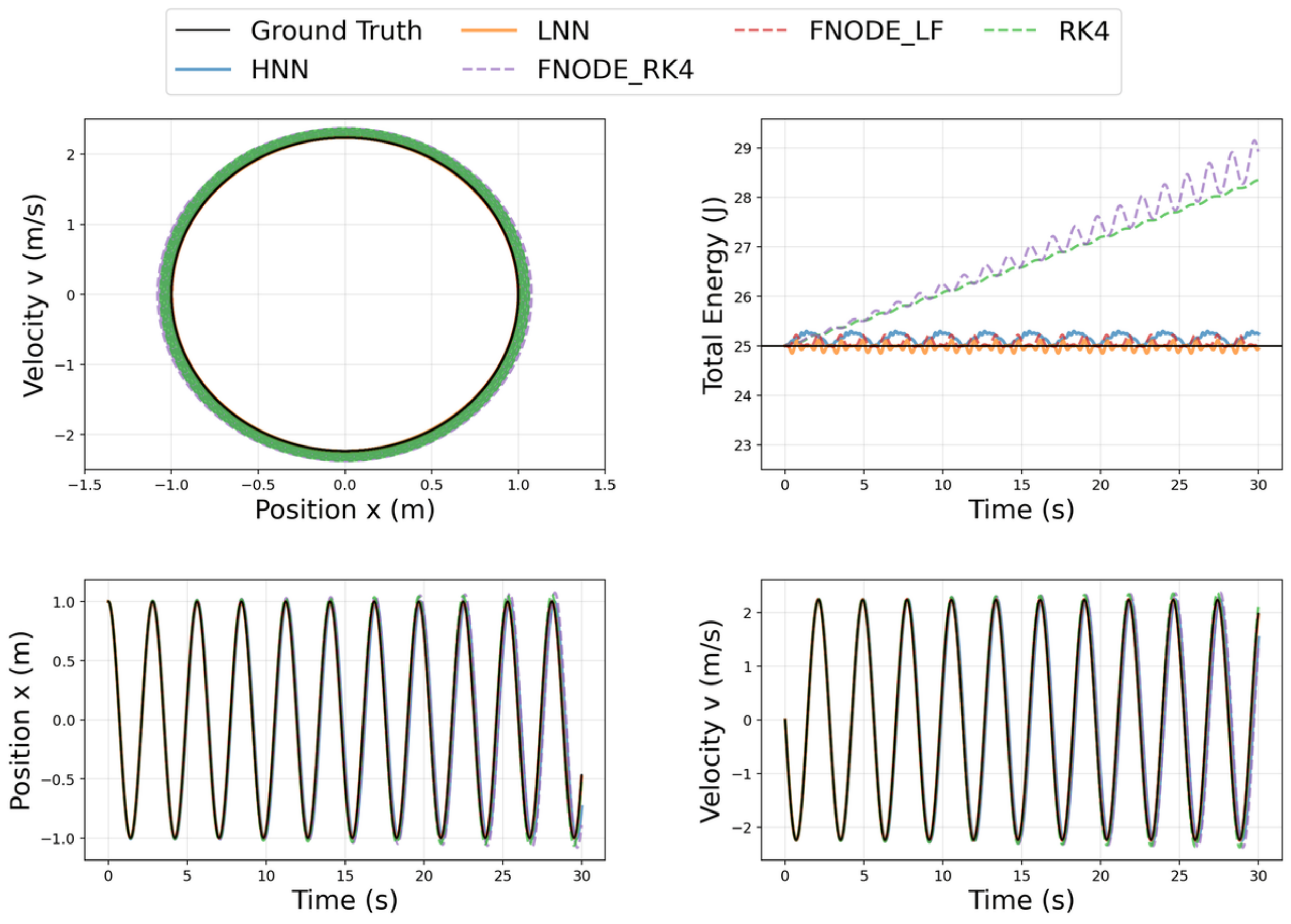}
	\caption{\updatedText{Phase-space trajectories and energy evolution for the single mass--spring system. The left panel shows the phase portrait ($v$ vs. $x$), while the right panel shows the total energy.
	The energy drift ratios are: FNODE\_RK4 ($15.70\%$), FNODE\_LF ($0.46\%$), RK4 ($13.39\%$), HNN ($0.98\%$), and LNN ($-0.24\%$).
	The mse for each model is provided: FNODE\_RK4 ($\epsilon=7.50\text{e-}2$), FNODE\_LF ($\epsilon=9.66\text{e-}5$), RK4 ($\epsilon=2.02\text{e-}3$), HNN ($\epsilon=5.51\text{e-}2$), and LNN ($\epsilon=2.27\text{e-}4$). The ground truth is provided by the analytical solution of the system.
	}}
	\label{fig:sms_state}
\end{figure}

\begin{table}[htbp]
	\centering
	\caption{\updatedText{Hyper-parameters for the single mass spring system.}}
	\label{tab:hyper_sms}
	{
	\begin{tabular}{@{}lccccc@{}}
		\toprule
		Hyper-parameters & \multicolumn{3}{c}{Model} \\
		\cmidrule(r){2-4}
		& FNODE & HNN & LNN \\ \midrule
		No. of hidden layers     & 2     & 2  & 2  \\
		No. of nodes per hidden layer &256  &256  &256   \\
		Max. epochs            &450        &30000  &400   \\
		Initial learning rate &1e-3 &  1e-3  & 1e-3 \\
		Learning rate decay &0.98 & 0.98  & 0.98 \\ 
		Activation function &tanh & tanh & tanh \\
		Loss function        &MSE   & MSE   & MSE   \\
		Optimizer    &Adam   & Adam   & Adam   \\  \bottomrule
	\end{tabular}
	}
\end{table}

\updatedText{Figure~\ref{fig:sms_state} compares long-term rollout behavior over the 30-second trajectory. In phase space, all methods remain close to the ground truth closed orbit, indicating that they capture the periodic motion. However, the energy trajectories highlight marked differences in stability. Specifically, structure-aware modeling and numerical integration, e.g., use of the Hamiltonian or Lagrangian structure or using a symplectic integrator (leap-frog in our implementation -- FNODE\_LF), mitigate energy drift, see Fig.~\ref{fig:sms_state}. HNN and LNN maintain near-constant energy throughout the rollout. In contrast, the RK4 integrator shows tangible energy drift, both when used directly on the equations of motion or in conjunction with the FNODE model (see FNODE\_RK4).
}

\subsection{Single-Mass-Spring-Damper System}
\label{sec:single_mass_spring_damper}
The equation of motion for the single-mass-spring-damper system is given by
\begin{equation}\label{eqn: smsd}
	\ddot{x} = -\tfrac{k}{m}x - \tfrac{d}{m}\dot{x} \;,
\end{equation}
where $m=10$ kg, $k=50$ N/m, and $d=2$ Ns/m. The system is initialized with $x(0)=1$ m, $\dot{x}(0)=0$, and solved by a Runge-Kutta 4th order (RK4) formula at $\Delta t=0.01$ s. A trajectory of 300 steps was used for training, with an additional 100 steps reserved for extrapolation. The model hyperparameters are listed in Table~\ref{tab:hyper_smsd}.

\begin{table}[htbp]
	\centering
	\caption{Hyper-parameters for the single mass-spring-damper system.}
	\label{tab:hyper_smsd}
	\begin{tabular}{@{}lccccc@{}}
		\toprule
		Hyper-parameters & \multicolumn{4}{c}{Model} \\
		\cmidrule(r){2-5}
		& FNODE  & MBD-NODE  &LSTM & FCNN \\ \midrule
		No. of hidden layers     & 2            & 2                            & 2    & 2   \\
		No. of nodes per hidden layer &256 & 256  & 256 & 256  \\
		
		Max. epochs            &450      & 300       & 300                     & 450                  \\
		Initial learning rate &1e-3 &  1e-3 &1e-3 & 1e-3 \\
		Learning rate decay &0.98 & 0.98  &0.98    &  0.98       \\ 
		Activation function &tanh & tanh &  Sigmoid,tanh &tanh\\
		Loss function        &MSE   & MSE   & MSE & MSE                    \\
		Optimizer              &Adam   & Adam & Adam & Adam                                       \\  \bottomrule
	\end{tabular}
\end{table}

\begin{figure}[htbp]
	\centering
	\includegraphics[width=12cm]{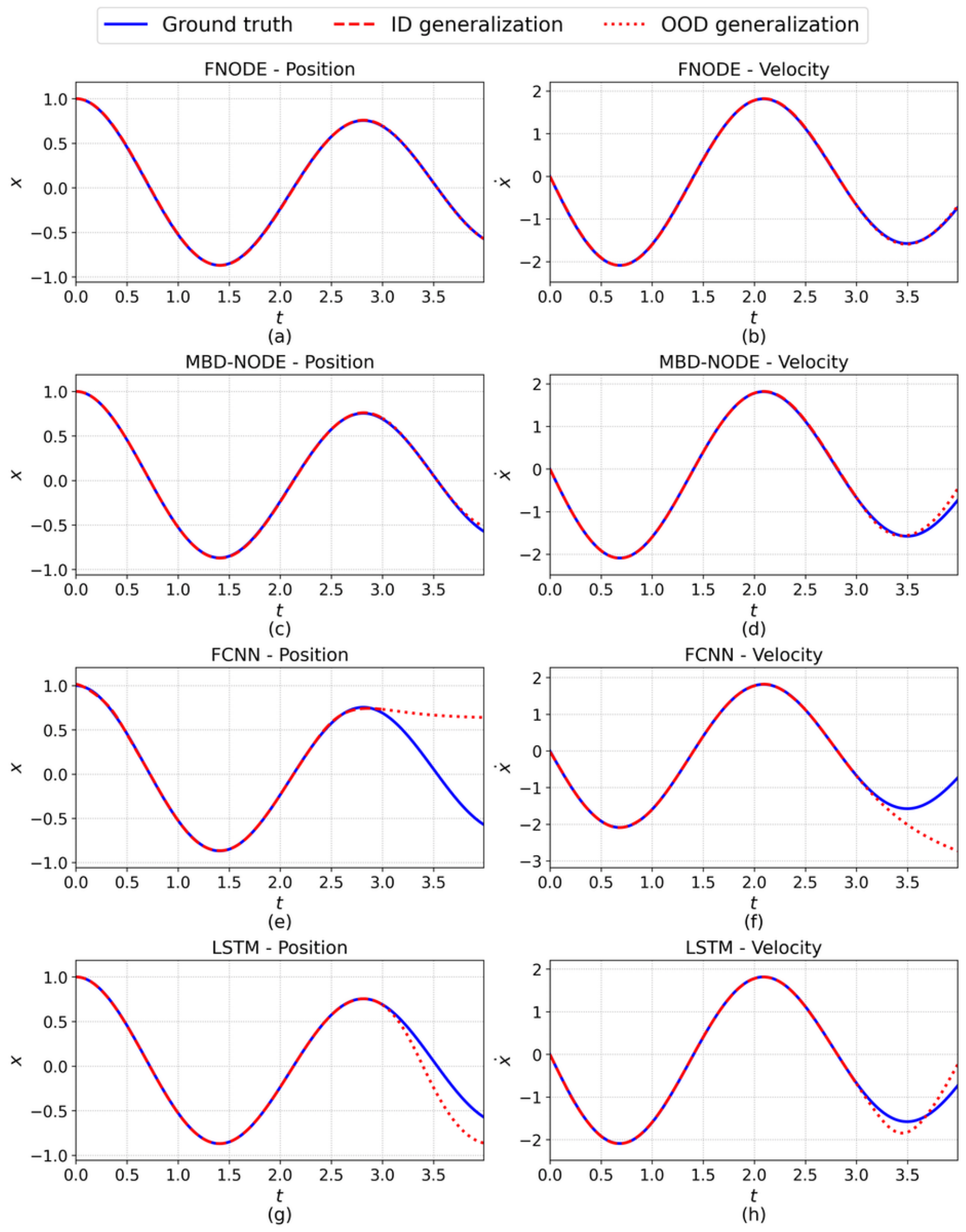}
	\caption{\updatedText{Temporal evolution of state variables for the single-mass-spring-damper system. ID stands for ``in distribution'' and OOD stands for ``out of distribution''. The left and right columns depict the displacement ($x$) and velocity ($\dot{x}$), respectively. Dashed lines indicate model predictions over the training interval ($t \in [0,3]$), while dotted lines represent extrapolated predictions on the test interval. The mean squared error (MSE) for each model is provided: FNODE ($\epsilon=7.6\text{e-}5$), MBD-NODE ($\epsilon=1.4\text{e-}3$), LSTM ($\epsilon=1.5\text{e-}2$), and FCNN ($\epsilon=1.6\text{e-}1$). The ground truth is provided by the analytical solution of the system.}}
	\label{fig:smsd_xvt}
\end{figure}

\begin{figure}[htbp]
	\centering
	\includegraphics[width=12cm]{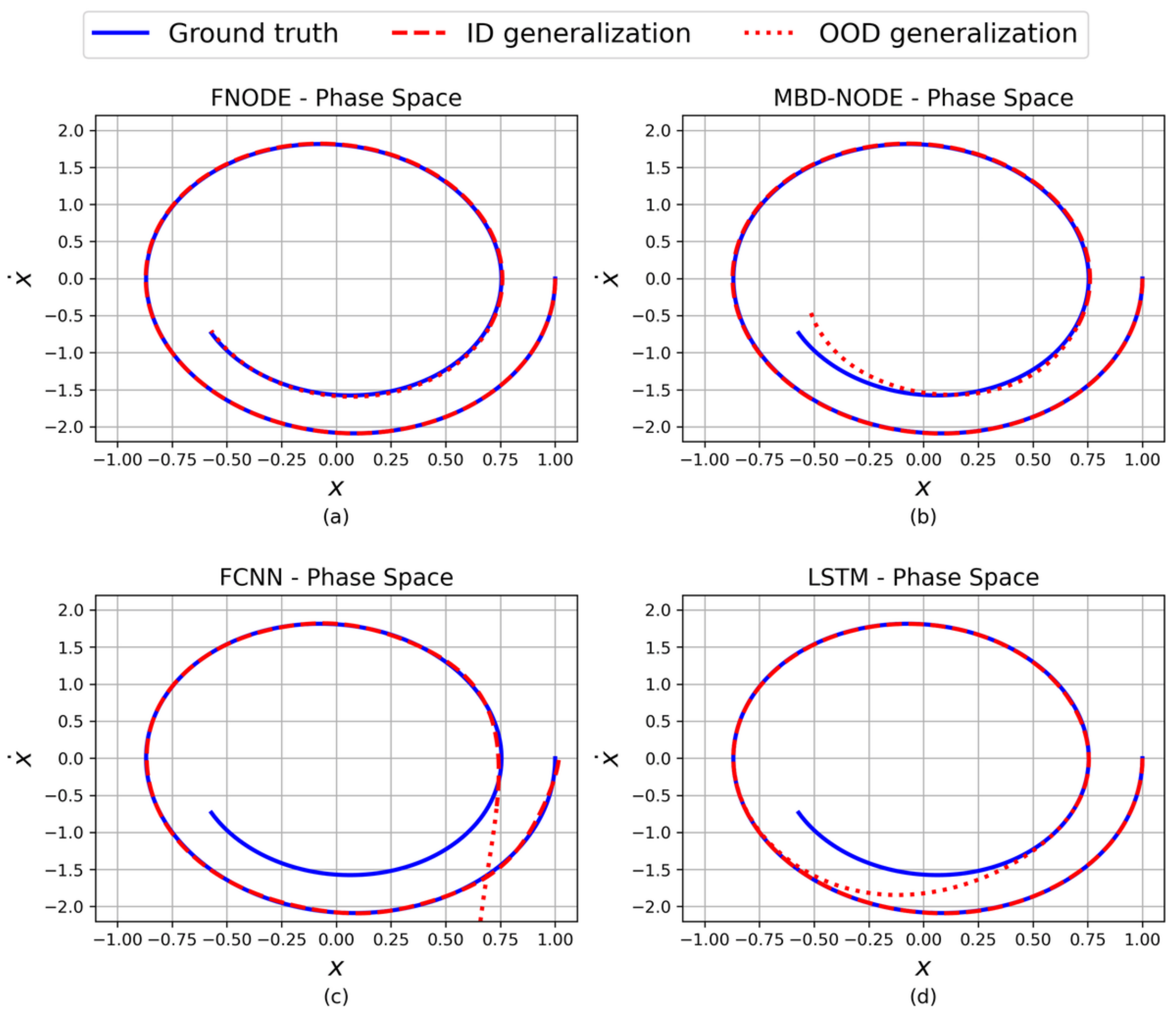}
	\caption{\updatedText{Comparative phase-space trajectories ($\dot{x}$ vs. $x$) for the single-mass-spring-damper system. Model performance on the training and test data is differentiated by dashed and dotted lines, respectively. The subplots correspond to the predictions of (a) FNODE, (b) MBD-NODE, (c) LSTM, and (d) FCNN.}}
	\label{fig:smsd_phase}
\end{figure}

Figure~\ref{fig:smsd_xvt} compares predicted displacement and velocity. 
All methods fit well within the training interval, but differences emerge in extrapolation. 
FNODE maintains close agreement with the ground truth 
while MBD-NODE degrades more quickly. 
LSTM fails to capture the dissipative behavior, 
and FCNN diverges entirely.
Phase-space trajectories in Figure~\ref{fig:smsd_phase} confirm FNODE's stable long-term prediction, 
in contrast to the drift or stagnation observed in the baselines.

\updatedSecond{We also investigate the sources of rollout error in FNODE on the single-mass-spring-damper system.
All errors are calculated over a fixed physical time horizon $T=2s$ by MSE.
Extract 15 samples of time step $\Delta t\in[5\times 10^{-4}, 1\times 10^{-2}]$ and define the uniform grid $t_n=n\Delta t$ for
$n=0,\dots,N$, where $N-1=\frac{T}{\Delta t}$.
Define $Z_n=Z(t_n)\in\mathbb{R}^{2d}$ as the accurate reference solution, and
let $\widetilde{Z}_n=\widetilde{Z}(t_n)\in\mathbb{R}^{2d}$ denote the system state (e.g., position and velocity). Let $f$ denote the ground-truth vector field $\ddot{z}=f(Z_n)$, and let $f_{\Theta}$ denote the learned vector field $\mathring{\ddot{z}}=f_{\Theta}(Z_n)$, see Eq.~(\ref{eqn: nn}).
}

\begin{figure}[htbp]
	\centering
	\includegraphics[width=12cm]{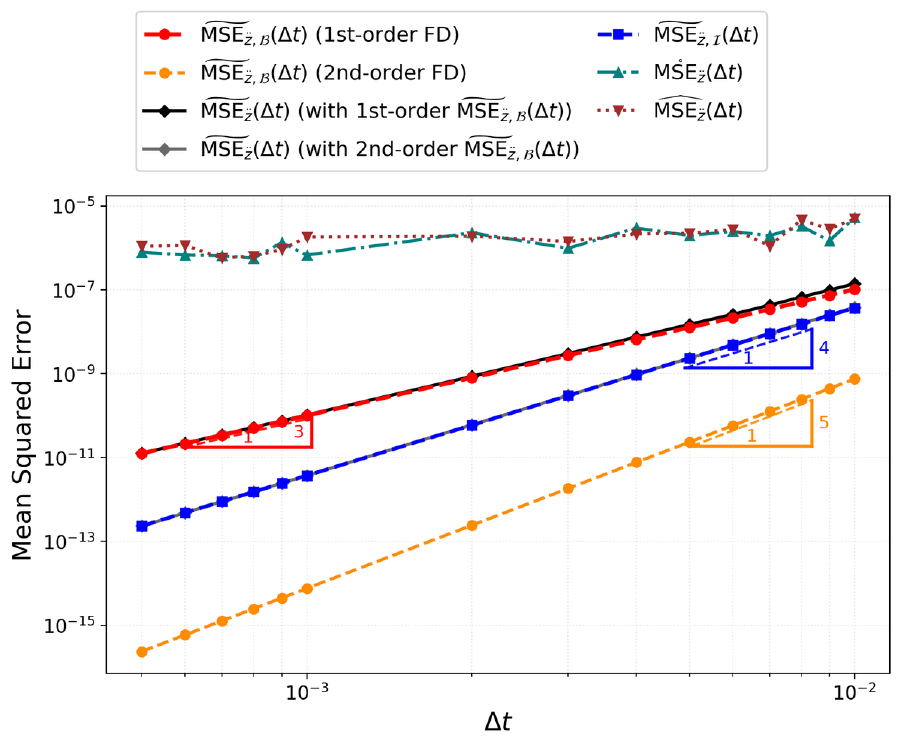}
	\caption{\updatedSecond{Log-log slope analysis for acceleration MSE.
	$\widetilde{\mathrm{MSE}}_{\ddot{z},\mathcal{B}}(\Delta t)$ (1st-order FD) is the MSE of boundary point acceleration with 1st-order finite difference,
	$\widetilde{\mathrm{MSE}}_{\ddot{z},\mathcal{B}}(\Delta t)$ (2nd-order FD) is the MSE of boundary point acceleration with 2nd-order central finite difference,
	$\widetilde{\mathrm{MSE}}_{\ddot{z},\mathcal{I}}(\Delta t)$ is the MSE of interior point acceleration with 2nd-order central finite difference, 
	$\widetilde{\mathrm{MSE}}_{\ddot{z}}(\Delta t)$ (with 1st-order $\widetilde{\mathrm{MSE}}_{\ddot{z},\mathcal{B}}(\Delta t)$) is the MSE of the overall acceleration in which the 
	boundary point acceleration is calculated by 1st-order finite difference,
	$\widetilde{\mathrm{MSE}}_{\ddot{z}}(\Delta t)$ (with 2nd-order $\widetilde{\mathrm{MSE}}_{\ddot{z},\mathcal{B}}(\Delta t)$) is the MSE of the overall acceleration in which the 
	boundary point acceleration is calculated by 2nd-order central finite difference,
	$\mathring{\mathrm{MSE}}_{\ddot{z}}(\Delta t)$ is the MSE of the overall acceleration predicted by the learned vector field $f_{\Theta}$ trained on the analytical acceleration solution,
	and $\widehat{\mathrm{MSE}}_{\ddot{z}}(\Delta t)$ is the MSE of the overall acceleration predicted by the learned vector field $f_{\Theta}$ trained on the finite difference acceleration result.
	The slope of each curve is also provided, and the ground truth is provided by the analytical acceleration solution of the system.
	}}
	\label{fig:accel_mse_decomposition}
\end{figure}

\updatedSecond{Figure~\ref{fig:accel_mse_decomposition} reports the decomposition of the acceleration MSE into boundary 
and interior contributions, together with the overall MSE. The log--log slope of each curve is also provided, and the 
ground truth is given by the analytical acceleration solution of the system. Consistent with the analysis in 
Section~\ref{sec:error-propagation}, the contribution with the lower FD order dominates the overall MSE. 
In particular, the boundary-point error (1st order FD) exhibits a slope close to $3$, 
matching the theoretical scaling $\mathcal{O}(\Delta t^{3})$ for the MSE of 1st order FD. 
The interior-point error has a slope close to $4$, consistent with the second-order central 
finite difference used for interior-point acceleration calculation. Accordingly, 
the overall MSE follows the scaling of the dominant term in the regime shown.
The same figure also compares the MSE of the learned vector field $f_{\Theta}$ trained on the analytical acceleration 
solution with that of $f_{\Theta}$ trained on the finite difference acceleration result. In both cases, the learned acceleration 
error is significantly larger than the FD error, and the overall MSE is dominated by the learned-field mismatch. 
This suggests that, in FNODE, the acceleration 	MSE is dominated by the error in the learned vector field.}

\begin{figure}[htbp]
	\centering
	\includegraphics[width=12cm]{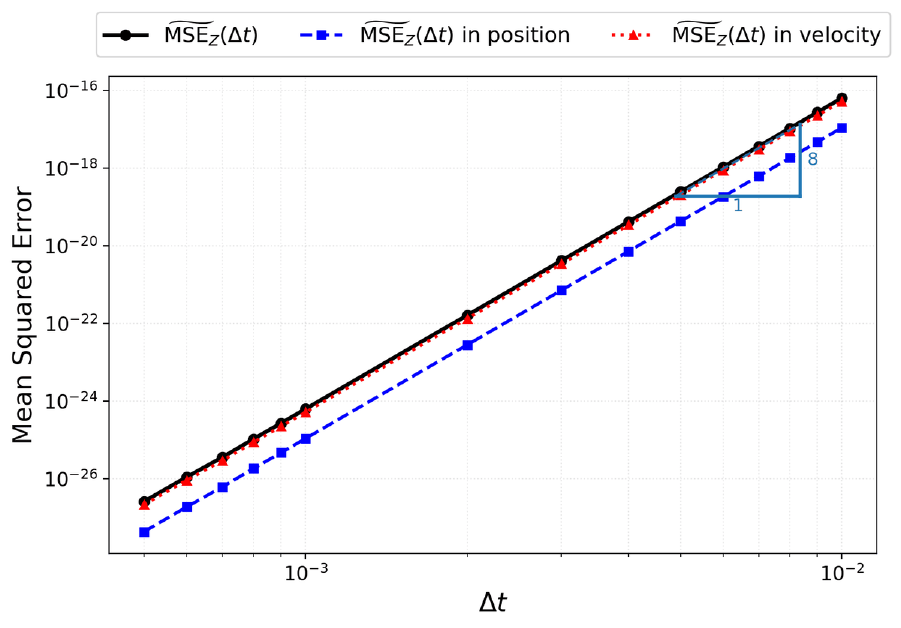}
	\caption{\updatedSecond{Log--log slope analysis for trajectory MSE of the RK4 integrator.}}
	\label{fig:traj_mse_slope}
\end{figure}

\updatedSecond{Figure~\ref{fig:traj_mse_slope} shows the log--log slope results for trajectory MSE of the RK4 integrator. 
The measured slope is close to $8$, which is consistent with the theoretical order of accuracy of RK4. 
This confirms that, when the learned vector field is accurate, the trajectory MSE is dominated by the integrator error.}

\subsection{Multiscale Triple-Mass-Spring-Damper System}\label{sec:triple_mass_spring_damper}

\begin{figure}[htbp]
	\centering
	\includegraphics[width=0.75\linewidth]{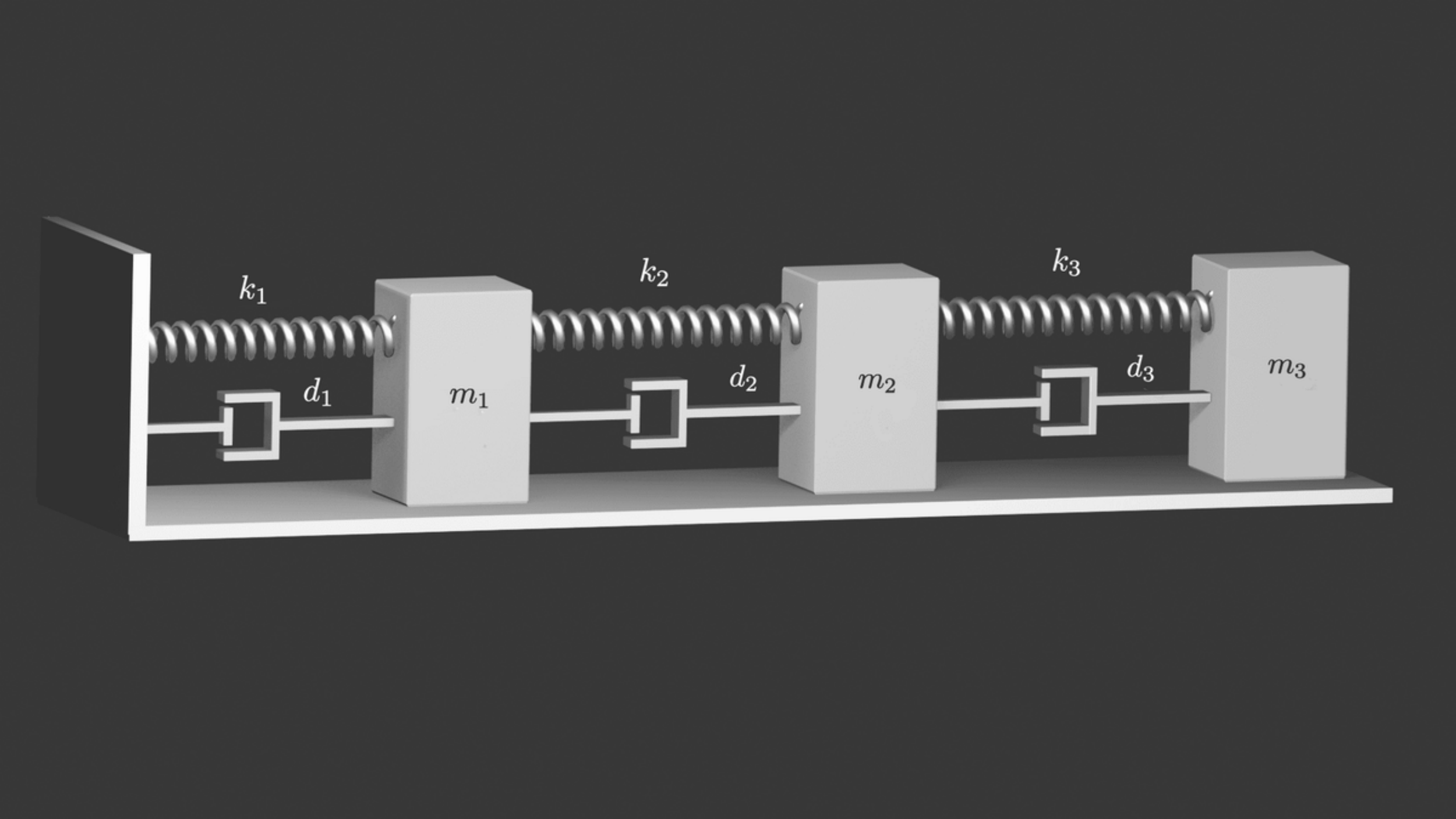}
	\caption{Triple mass-spring-damper system. The setup is similar to the single mass-spring-damper system, except for the addition of two more masses, springs, and dampers. \updatedText{The system admits an analytical solution that was used for accuracy studies.}}
	\label{fig:tmsd_figure}
\end{figure}

We next consider a three-mass-spring-damper system with strong scale separation (Figure~\ref{fig:tmsd_figure}). The masses are $m_1=100$ kg, $m_2=10$ kg, and $m_3=1$ kg. All springs have stiffness $k_i=50$ N/m and all dampers $d_i=2$ Ns/m. The governing equations are
\begin{equation}
	\begin{aligned}
		\ddot{x}_1 &= -\tfrac{k_1}{m_1} x_1 - \tfrac{d_1}{m_1} \dot{x}_1  + \tfrac{k_2}{m_1} (x_2 - x_1) + \tfrac{d_2}{m_1} (\dot{x}_2 - \dot{x}_1) \\
		\ddot{x}_2 &= -\tfrac{k_2}{m_2} (x_2 - x_1) - \tfrac{d_2}{m_2} (\dot{x}_2 - \dot{x}_1) + \tfrac{k_3}{m_2} (x_3 - x_2) + \tfrac{d_3}{m_2} (\dot{x}_3 - \dot{x}_2) \\
		\ddot{x}_3 &= -\tfrac{k_3}{m_3} (x_3 - x_2) - \tfrac{d_3}{m_3} (\dot{x}_3 - \dot{x}_2)
	\end{aligned} \; .
\end{equation}

The system was simulated with RK4 at a fixed step of $\Delta t=0.01$ s. 
Training trajectories consisted of 300 steps from the initial state $x_1=1$, $x_2=2$, $x_3=3$, $\dot{x}_1=\dot{x}_2=\dot{x}_3=0$ (SI units). 
Performance was then evaluated by extrapolation for an additional 100 steps. Model hyperparameters are listed in Table~\ref{tab:hyper_tmsd}.
\begin{table}[htbp]
	\centering
	\caption{Hyper-parameters for the triple-mass-spring-damper system.}
	\label{tab:hyper_tmsd}
	\begin{tabular}{@{}lccccc@{}}
		\toprule
		Hyper-parameters & \multicolumn{4}{c}{Model} \\
		\cmidrule(r){2-5}
		& FNODE  & MBD-NODE  &LSTM & FCNN \\ \midrule
		No. of hidden layers     & 2            & 2             & 2    & 2   \\
		No. of nodes per hidden layer &256 & 256  & 256 & 256  \\
		
		Max. epochs            &450      & 300       & 300                     & 450                  \\
		Initial learning rate &1e-3 &  1e-3 &1e-3 & 1e-3 \\
		Learning rate decay &0.98 & 0.98  &0.98    & 0.98       \\ 
		Activation function &tanh & tanh & Sigmoid,tanh &tanh\\
		Loss function        &MSE   & MSE   & MSE & MSE                    \\
		Optimizer              &Adam   & Adam & Adam & Adam                                       \\  \bottomrule
	\end{tabular}
\end{table}

\begin{figure}[htbp]
	\centering
	\includegraphics[width=12cm]{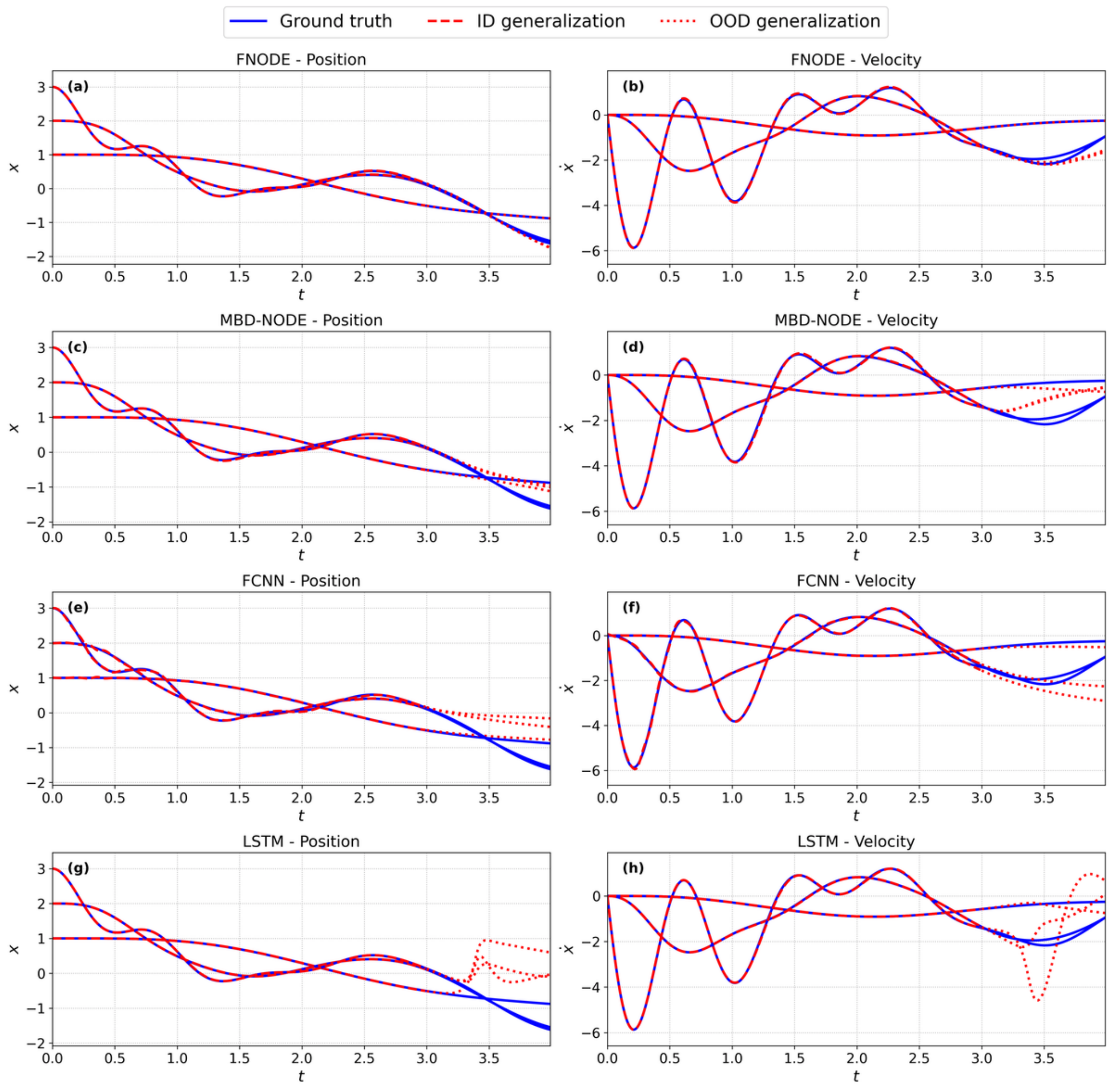}
	\caption{\updatedText{Predicted temporal evolution for the triple-mass-spring-damper benchmark. The left column presents the displacement ($x$) and the right column presents the velocity ($\dot{x}$) for each of the three masses. Dashed and dotted lines denote predictions on the training ($t \in [0,3]$) and test sets, respectively. The associated MSEs are: FNODE ($\epsilon=6.1\text{e-}3$), MBD-NODE ($\epsilon=5.4\text{e-}2$), LSTM ($\epsilon=2.5\text{e-}1$), and FCNN ($\epsilon=8.9\text{e-}2$).}}
	\label{fig:tmsd_xvt}
\end{figure}

Figure~\ref{fig:tmsd_xvt} shows the trajectories of $x_1,x_2,x_3$ and $\dot{x}_1,\dot{x}_2,\dot{x}_3$ during training and testing. All models fit the training data, though FCNN exhibits minor \updatedText{spurious} oscillations. In extrapolation ($t>3$), FNODE achieves the best agreement with ground truth ($\epsilon=7.4\text{e-}4$). MBD-NODE remains reasonable but drifts due to solver error accumulation. LSTM largely repeats training patterns and fails to generalize, while FCNN diverges under multiscale dynamics.

\begin{figure}[htbp]
	\centering
	\includegraphics[width=12cm]{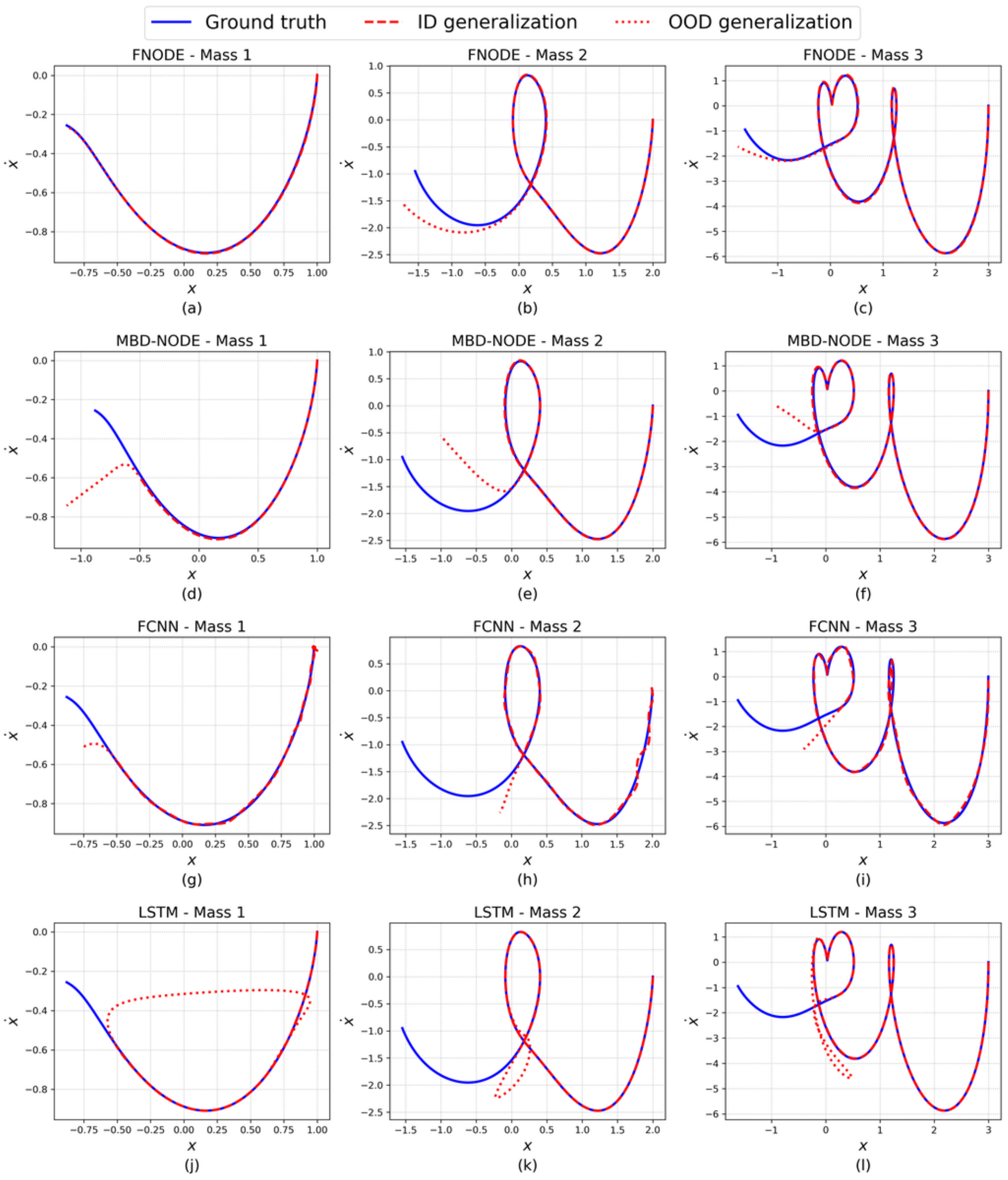}
	\caption{\updatedText{Phase-space trajectories for the three constituent masses of the triple-mass-spring-damper system. Each column corresponds to a different mass. Model performance on the training and test datasets is indicated by dashed and dotted lines, respectively. The rows correspond to the results from (a-c) FNODE, (d-f) MBD-NODE, (g-i) LSTM, and (j-l) FCNN.}}
	\label{fig:tmsd_phase}
\end{figure}

\updatedText{Figure~\ref{fig:tmsd_phase} shows phase-space trajectories on the test set. FNODE tracks the dynamics well -- the small discrepancies for the lighter masses are consistent with increased sensitivity to local inaccuracies in the learned acceleration field and their subsequent accumulation during rollout, as discussed in Section~\ref{sec:error-propagation}}. MBD-NODE performs poorly for Mass~1, where low variance in the signal reduces its contribution to the loss. LSTM largely repeats training patterns and fails to generalize, while FCNN is inaccurate both in-distribution and out-of-distribution.

\FloatBarrier

\subsection{Double Pendulum}
\label{sec:double_pendulum}
We next evaluate FNODE on a classic example of a chaotic system -- the double pendulum. The system is described by the Hamiltonian formulation (see \ref{app:double_pendulum} for details). We used $L_1=L_2=1$ m, $m_1=m_2=1$ kg, and $g=9.81$ m/s$^2$. The initial state was $\theta_1=3\pi/7$, $\theta_2=3\pi/4$, and $\dot{\theta}_1=\dot{\theta}_2=0$. Trajectories were generated using RK4 with $\Delta t=0.01$ s for 300 training steps and 100 extrapolation steps. The model hyperparameters are listed in Table~\ref{tab:hyper_dp}.

\begin{table}[h]
	\centering
	\caption{Hyper-parameters for the double pendulum system}
	\label{tab:hyper_dp}
	\begin{tabular}{@{}lccccc@{}}
		\toprule
		Hyper-parameters & \multicolumn{4}{c}{Model} \\
		\cmidrule(r){2-5}
		&FNODE & MBD-NODE          &LSTM & FCNN \\ \midrule
		No. of hidden layers    & 2 & 2          & 2          & 2                      \\
		No. of nodes per hidden layer & 256 & 256 & 256 & 256  \\
		
		Max. epochs             & 450     & 300       & 300                     & 450                  \\
		Initial learning rate  &  3e-3 &  1e-3  &1e-3 & 3e-3 \\
		Learning rate decay & 0.98 & 0.98  &0.98    &  0.98       \\ 
		Activation function & tanh & tanh & \updatedText{Sigmoid,tanh} &tanh\\
		Loss function          & MSE  & MSE & MSE& MSE                   \\
		Optimizer              & Adam      & Adam     & Adam& Adam                                  \\  \bottomrule
	\end{tabular}
\end{table}

\begin{figure}
	\centering
	\includegraphics[width=0.75\linewidth]{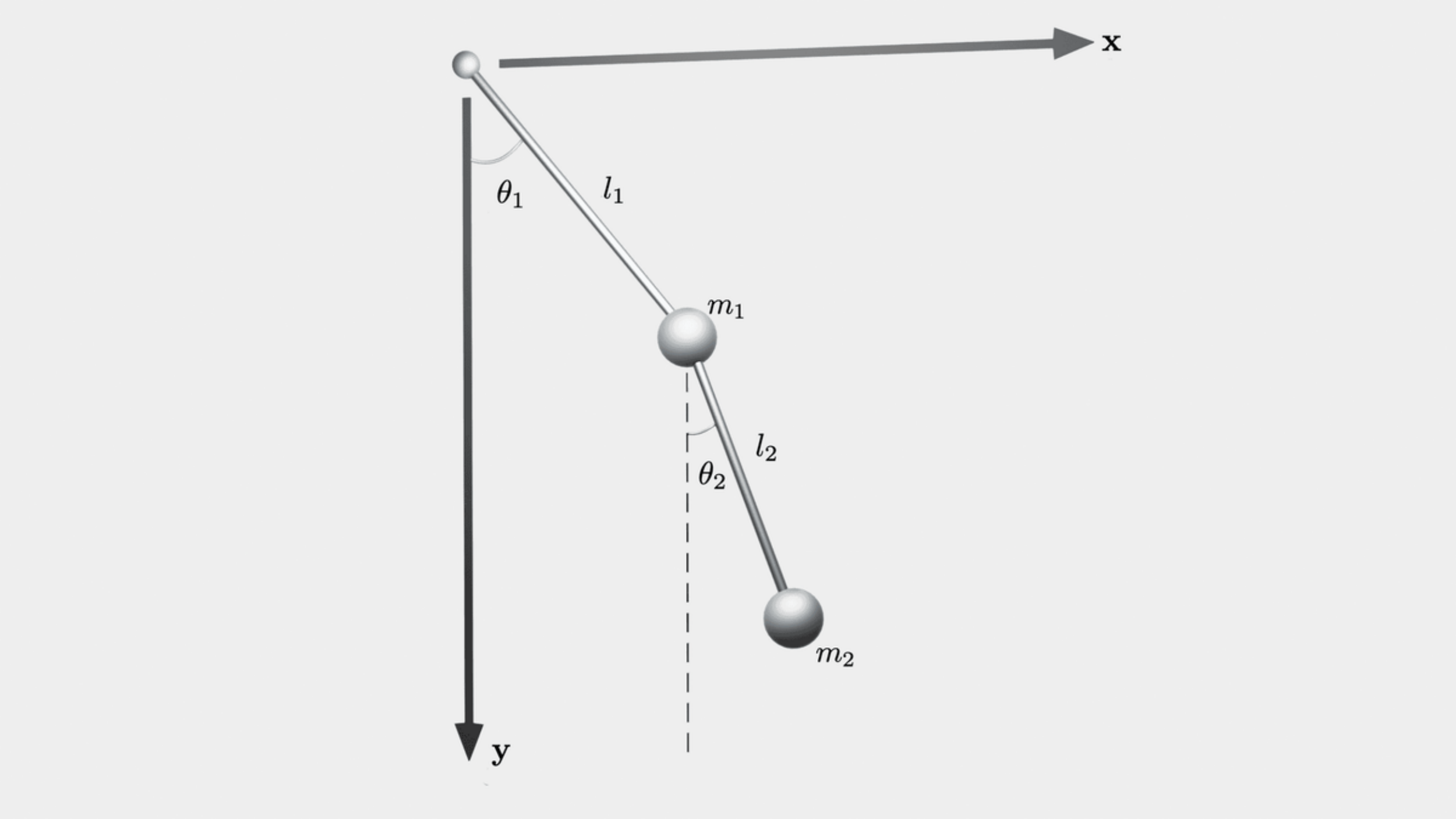}
	\caption{Schematic of the double pendulum.}
	\label{fig:double_pendulum}
\end{figure}

\begin{figure}[htbp]
	\centering
	\includegraphics[width=12cm]{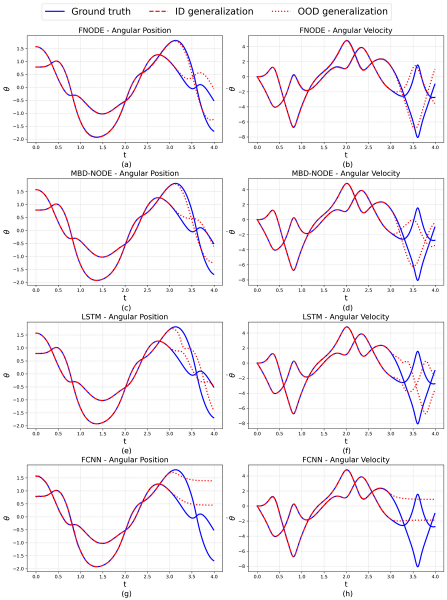}
	\caption{Dynamic response predictions for the double pendulum system. The left and right columns show the temporal evolution of angular position ($\theta$) and angular velocity ($\dot{\theta}$), respectively. Dashed lines represent the fit to the training data ($t \in [0,3]$), while dotted lines show the extrapolation on test data. The reported MSE values are: FNODE ($\epsilon=1.4\text{e-}1$), MBD-NODE ($\epsilon=2.3\text{e-}1$), LSTM ($\epsilon=7.6\text{e-}1$), and FCNN ($\epsilon=1.9\text{e}0$).}
	\label{fig:dp_xvt}
\end{figure}

Figure~\ref{fig:dp_xvt} shows the double pendulum trajectories. All models fit the training data, but diverge in extrapolation. FNODE, despite error accumulation in the ODE solver, achieves the best accuracy ($\epsilon=1.4\text{e-}1$). 

\begin{figure}[htbp]
	\centering
	\includegraphics[width=12cm]{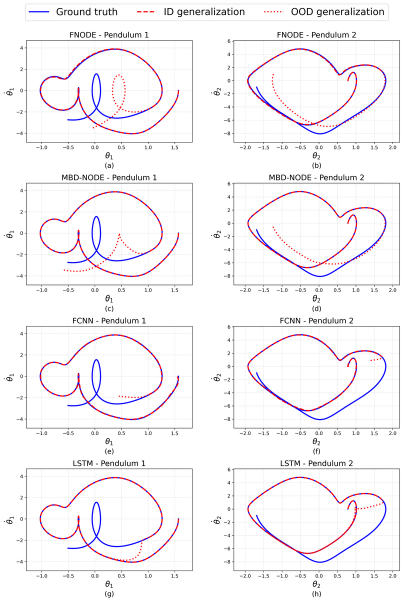}
	\caption{Phase-space of double pendulum system. The left column corresponds to the first mass and the right to the second. Dashed lines indicate the trajectory over the training interval, and dotted lines show the predicted trajectory over the test interval. The models evaluated are (a,b) FNODE, (c,d) MBD-NODE, (e,f) LSTM, and (g,h) FCNN.}
	\label{fig:dp_phase}
\end{figure}

Phase-space plots in Figure~\ref{fig:dp_phase} confirm this trend: FNODE reproduces the main chaotic patterns, with only minor discrepancies. MBD-NODE loses accuracy in testing, LSTM largely repeats training patterns without generalization, and FCNN fails to capture the dynamics outside the training regime.

\updatedText{For the double pendulum, which exhibits chaotic behavior in the regime considered here, ``long-term'' pointwise trajectory accuracy must be interpreted relative to the system's intrinsic predictability horizon. In chaotic dynamics, small initial-condition or modeling errors often grow approximately exponentially (on average) at a rate set by the largest Lyapunov exponent $\lambda_{\max}$, motivating the Lyapunov time scale $T_L := \tfrac{1}{\lambda_{\max}}$. Beyond horizons on the order of a few $T_L$, pointwise trajectory comparisons become increasingly stringent because even very small discrepancies are rapidly amplified by the dynamics. The rollout horizons reported in this section should therefore be interpreted as finite-horizon predictions (short to intermediate horizons) rather than claims of sustained pointwise accuracy arbitrarily far into the chaotic regime. Estimating $\lambda_{\max}$ for the specific operating regime (and reporting horizons in normalized units $t/T_L$) is an important direction for future work. Finally, even when long-horizon prediction is inherently challenging, FNODE (and related learned dynamics models) can still be useful in downstream tasks that rely on short-horizon predictive accuracy, e.g., when the prediction horizon of interest is short relative to $T_L$.}

\FloatBarrier
\subsection{\updatedText{Slider Crank with Coulomb friction}}
\label{sec:slider_crank}




\begin{figure}
	\centering
	\includegraphics[width=0.75\linewidth]{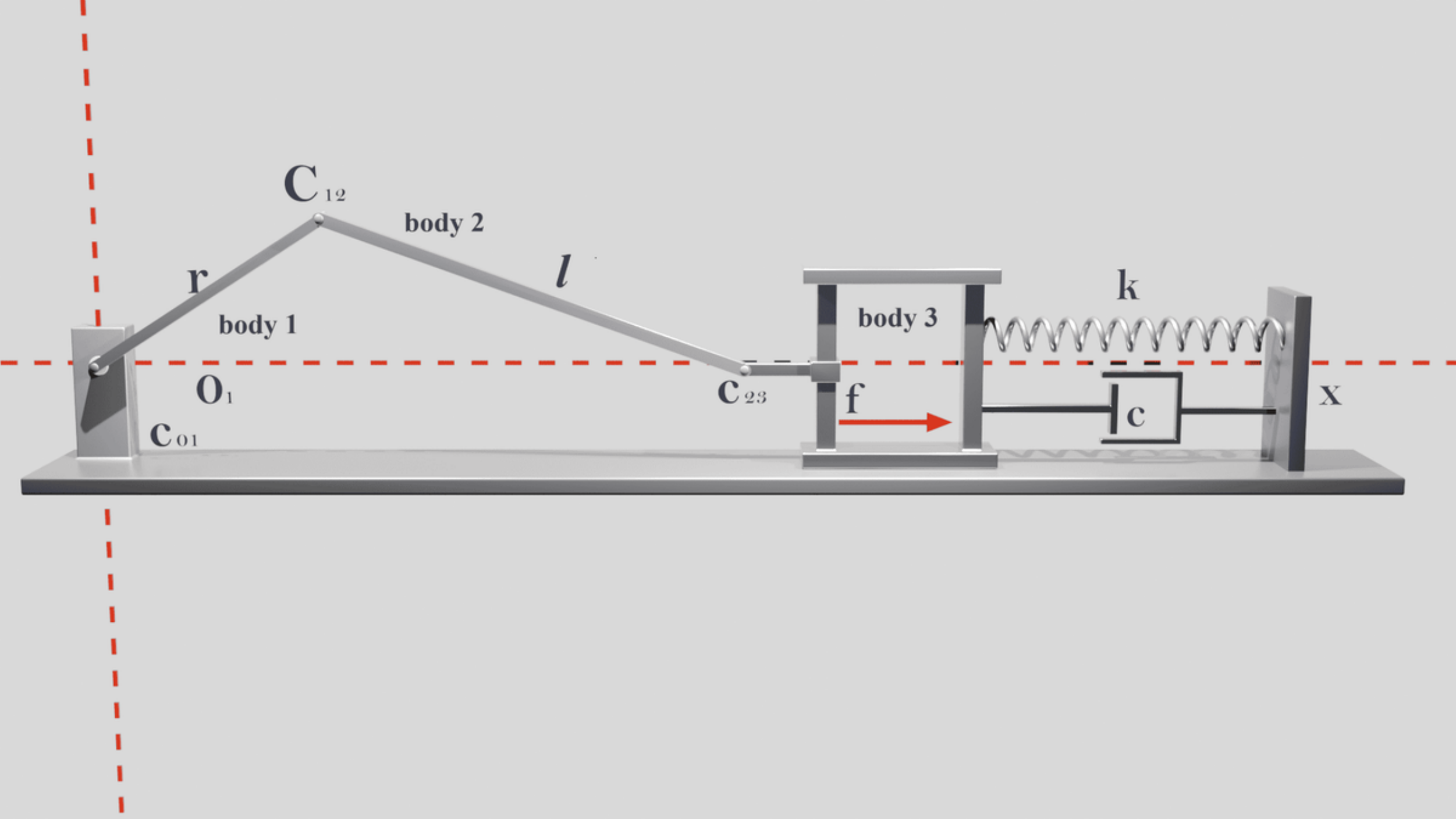}
	\caption{\updatedText{Slider-crank mechanism with motor torque $\tau_1$, rotational dampers at all joints, and slider friction. The friction force is shown in red, see \ref{sec:slider-crank-mechanism}, Eq.~(\ref{eqn:F_e3_with_friction}), for its expression. It acts between body 3 and ground.}}
	\label{fig:slider_crank}
\end{figure}

\updatedText{Next, FNODE's performance is evaluated on a constrained slider-crank mechanism (Figure~\ref{fig:slider_crank}) under both frictionless and frictional conditions. The mechanism consists of three rigid bodies with a slider connected to a spring and damper, forming a constrained three-body problem. The generalized coordinates are $q=(x_1, y_1, \theta_1, x_2, y_2, \theta_2, x_3, y_3, \theta_3)$; see \ref{sec:slider-crank-mechanism} for details. The mechanism is defined by the following three bodies:}

\begin{enumerate}
	\item A crank (body 1) with mass $m_1=1$ kg and moment of inertia $I_1=0.1 \text{ kg} \cdot \text{m}^2$ is connected to the ground by a revolute joint. Its length is $l=2$ m. 
	\item A connecting rod (body 2) with mass $m_2=1$ kg and moment of inertia $I_2=0.1 \text{ kg} \cdot \text{m}^2$ is attached to the crank via a revolute joint. Its length is $r=1$ m.
	\item A slider (body 3) with mass $m_3=1$ kg and moment of inertia $I_3=0.1 \text{ kg} \cdot \text{m}^2$ is joined to the connecting rod by a revolute joint. It is also connected to a fixed wall via a spring with a torsional constant of $k=1$ Nm/rad. There is friction between the slider and the ground, which is modeled as Coulomb friction with a coefficient of $\mu$.
\end{enumerate}

\updatedText{FNODE learns a mapping from the full 18-dimensional state space $(q,\dot{q})$ to accelerations $\ddot{q}$. Since the mechanism has only one DOF, the dynamics can be represented compactly with the minimal coordinates $(\theta_1,\dot{\theta}_1)$, with all other states dependent. Following the coordinate-partitioning approach \cite{Wehage82}, once FNODE produces the acceleration of the crank, numerical integration is used to compute the orientation and the angular velocity of the crank, while the kinematic constraint equations are used to recover the dependent generalized coordinates. The model hyperparameters are summarized in Table~\ref{tab:hyper_sc}.}

\begin{table}[h]
	\centering
	\caption{Hyper-parameters for the slider-crank mechanism.}
	\label{tab:hyper_sc}
	\begin{tabular}{@{}lccccc@{}}
		\toprule
		Hyper-parameters & \multicolumn{4}{c}{ Model} \\
		\cmidrule(r){2-5}
		&  FNODE &  MBD-NODE          & LSTM &  FCNN\\ \midrule
		No. of hidden layers     &  \updatedText{3} & \updatedText{3}  & \updatedText{3}  & \updatedText{3}      \\
		No. of nodes per hidden layer & 256  & 256  & 256  & 256   \\
		Max. epochs  &  450   &  300   & 300   & 450    \\
		Initial learning rate & \updatedText{5e-4} & 1e-3 &1e-3  & 1e-3 \\
		Learning rate decay &  \updatedText{0.99} &  0.98  &0.98    &  0.98   \\ 
		Activation function &  tanh &  tanh & Sigmoid,tanh  &tanh \\
		Loss function     & MSE  & MSE &  MSE &  MSE                   \\
		Optimizer     &  Adam    &  Adam     & Adam & Adam                                   \\  \bottomrule
	\end{tabular}
\end{table}

\subsubsection{Non-friction Case} 
\begin{figure}[htbp]
	\centering
	\includegraphics[width=12cm]{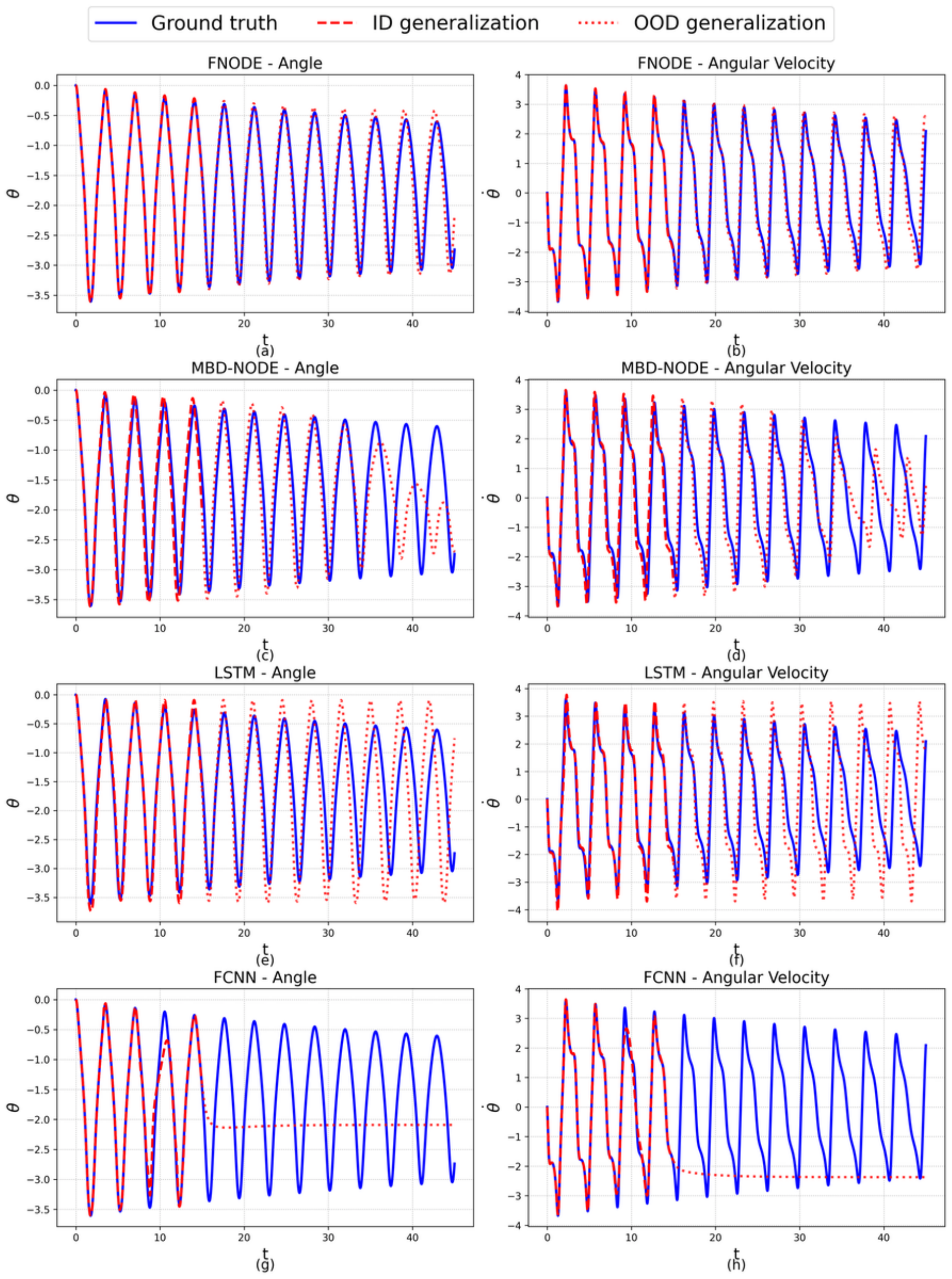}
	\caption{\updatedText{Long-term prediction results for the frictionless slider-crank's independent coordinate ($\theta_1$). The left column presents angular position versus time; the right column presents angular velocity versus time. Model performance on the training interval ($t \in [0,15]$) is shown with dashed lines, while extrapolated performance on the test interval is shown with dotted lines. The corresponding MSEs are: FNODE ($\epsilon=1.7\text{e-}1$), MBD-NODE ($\epsilon=8.7\text{e-}1$), LSTM ($\epsilon=1.4\text{e}0$), and FCNN ($\epsilon=3.1\text{e}0$).}}
	\label{fig:sc_xvt}
\end{figure}

\begin{figure}[htbp]
	\centering
	\includegraphics[width=12cm]{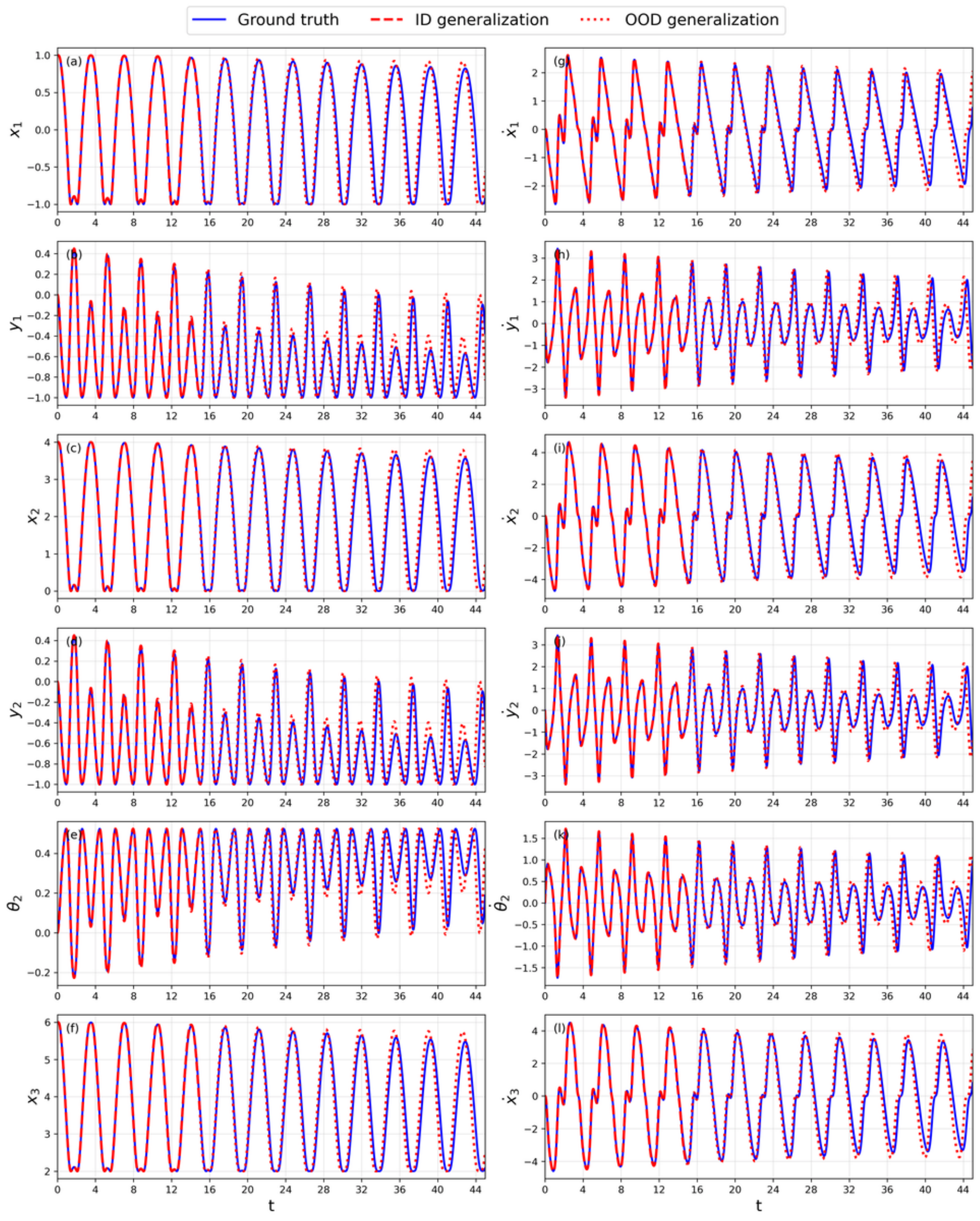}
	\caption{\updatedText{Temporal evolution of the dependent state variables for the slider-crank mechanism, as reconstructed from the FNODE model's prediction of the minimal coordinates.}}
	\label{fig:sc_all_states_fnode}
\end{figure}

\begin{figure}[htbp]
	\centering
	\includegraphics[width=12cm]{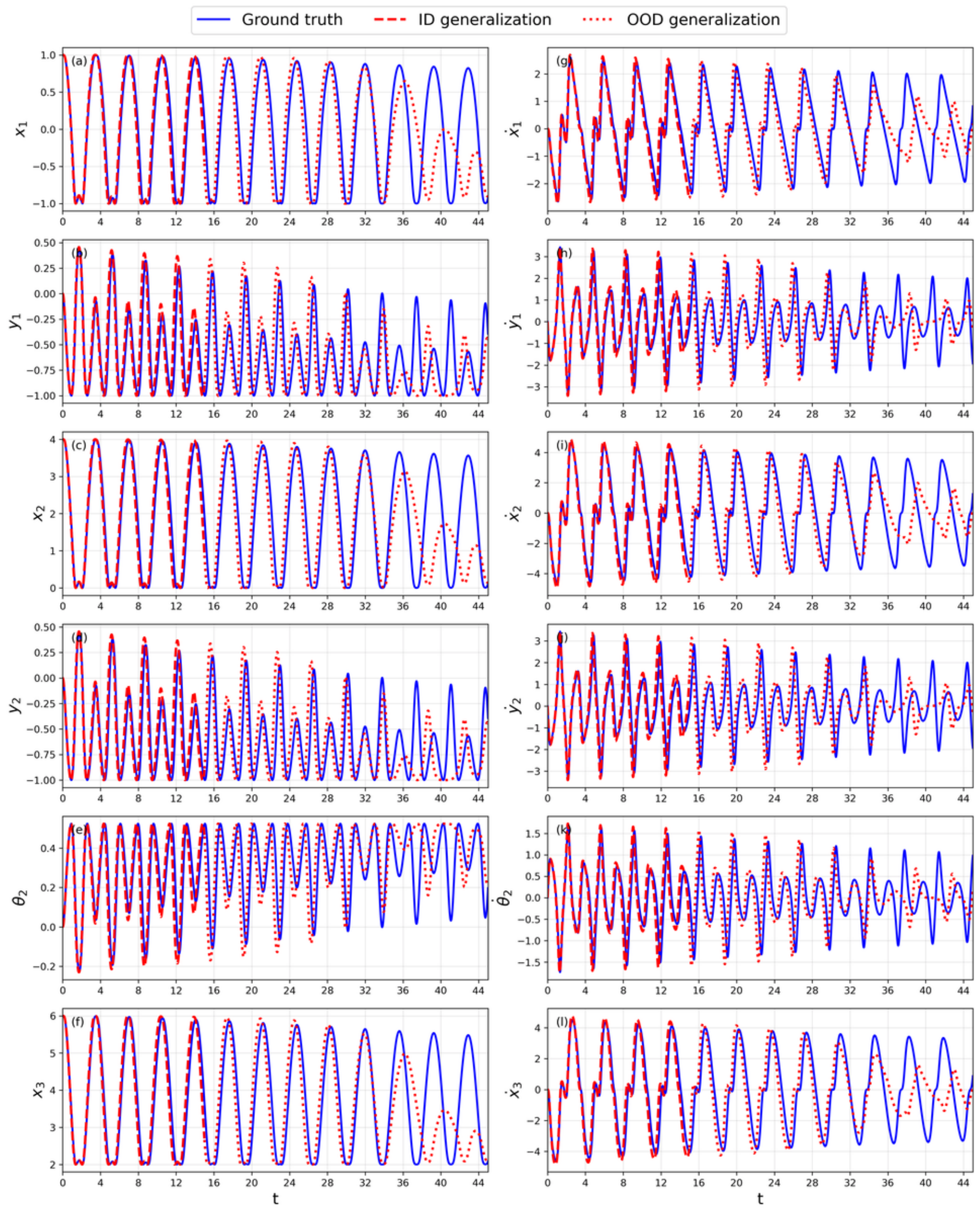}
	\caption{\updatedText{Temporal evolution of the dependent state variables for the slider-crank mechanism, as reconstructed from the MBD-NODE model's prediction of the minimal coordinates.}}
	\label{fig:sc_all_states_mbdnode}
\end{figure}

\begin{figure}[htbp]
	\centering
	\includegraphics[width=12cm]{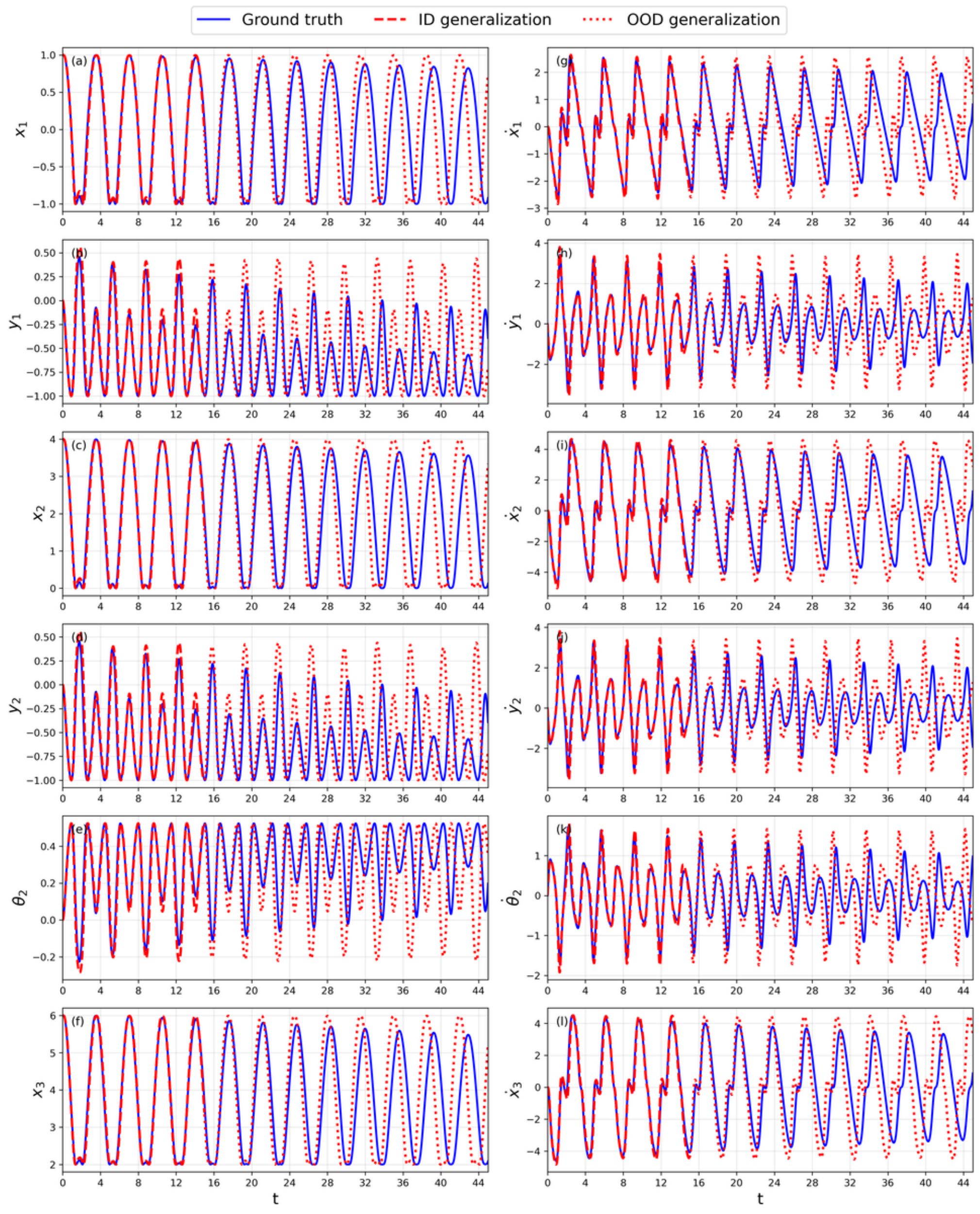}
	\caption{\updatedText{Temporal evolution of the dependent state variables for the slider-crank mechanism, as reconstructed from the LSTM model's prediction of the minimal coordinates.}}
	\label{fig:sc_all_states_lstm}
\end{figure}

\begin{figure}[htbp]
	\centering
	\includegraphics[width=12cm]{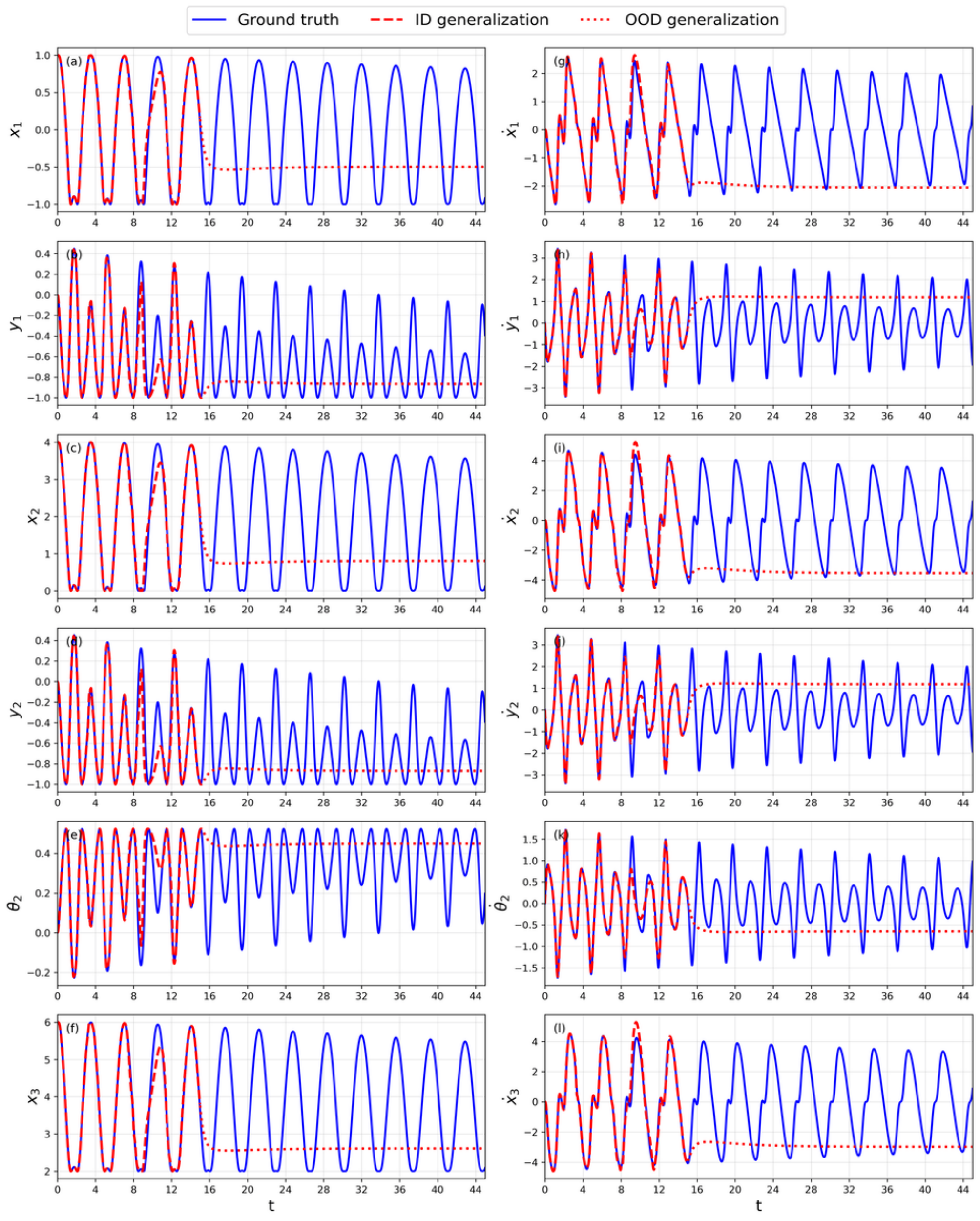}
	\caption{\updatedText{Temporal evolution of the dependent state variables for the slider-crank mechanism, as reconstructed from the FCNN model's prediction of the minimal coordinates.}}
	\label{fig:sc_all_states_fcnn}
\end{figure}

Figure~\ref{fig:sc_xvt} shows the slider-crank responses. 
All models fit the training data, but diverge in extrapolation. 
FNODE maintains accurate long-term predictions. 
MBD-NODE, while structurally similar to FNODE, is less stable and incurs higher computational cost due to its solver-based training. 
In contrast, LSTM and FCNN fail to generalize beyond the training window. 
Figures~\ref{fig:sc_all_states_fnode}--\ref{fig:sc_all_states_fcnn} illustrate the remaining state variables, which inherit errors from the minimal coordinates $(\theta_1,\dot{\theta}_1)$.

\newpage

\newpage

\subsubsection{\updatedText{Friction Case}}
\updatedText{In this experiment, the friction coefficient $\mu$ is added as an input to the FNODE model to test the interpolation in the parameter space. The friction coefficient is sampled from $[0.0,0.6]$ uniformly for training data generation. For testing, we use interpolated friction coefficients to test the generalization ability of FNODE in parameter space.}

\begin{figure}[htbp]
	\centering
	\includegraphics[width=12cm]{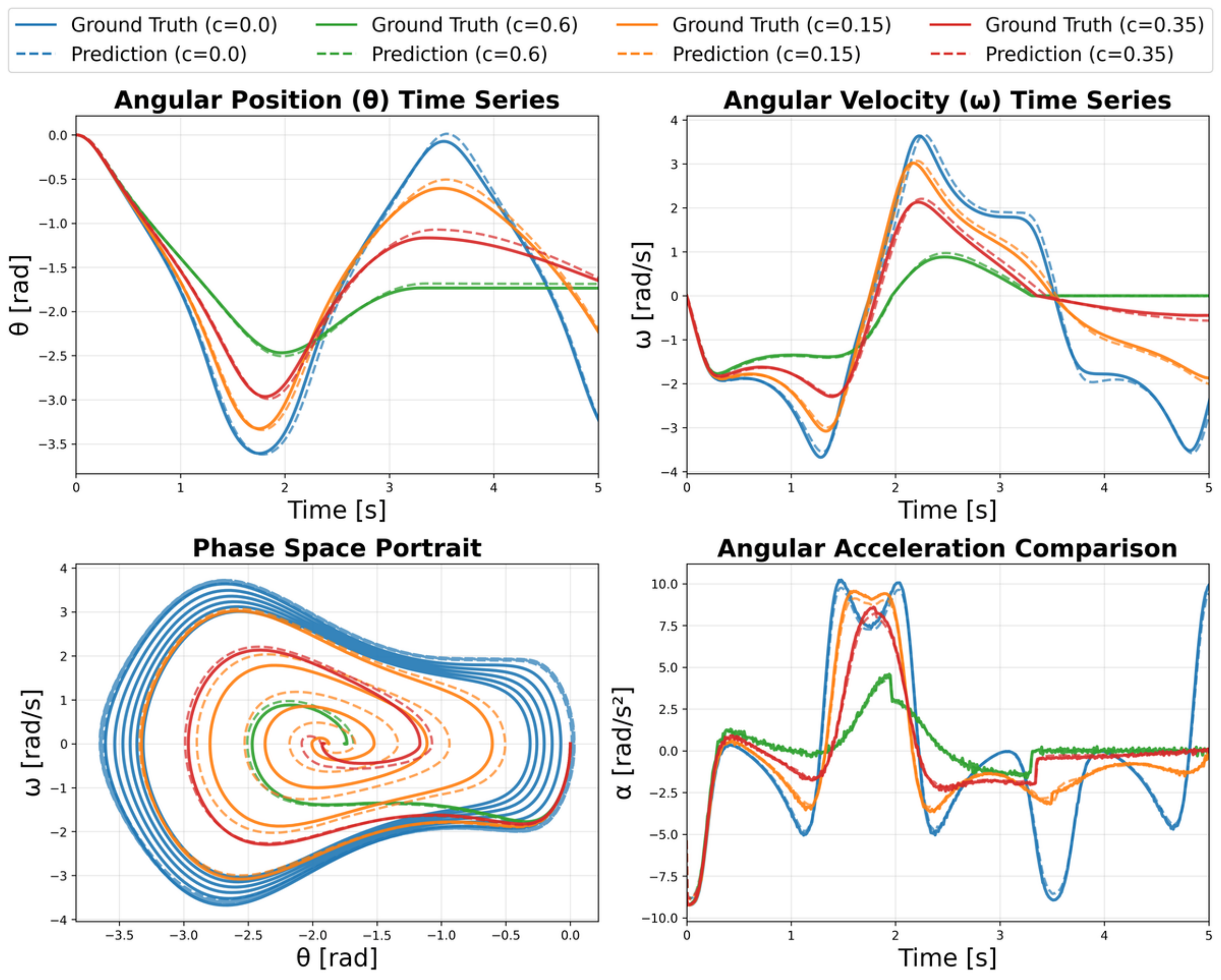}
	\caption{\updatedText{Trajectories of the slider-crank mechanism under varying friction coefficients, as predicted by the FNODE model. The model was trained on friction coefficients sampled uniformly from $[0.0,0.6]$ and tested on interpolated values.
	In this figure, c = 0.0 and c = 0.6 are from the training set, while c = 0.15, 0.35 are interpolated test cases.}}
	\label{fig:sc_phase_space}
\end{figure}

\updatedText{Figure~\ref{fig:sc_phase_space} shows FNODE-predicted phase-space trajectories of the slider-crank mechanism for different friction coefficients. The FNODE model was trained on coefficients sampled uniformly from $[0.0, 0.6]$ and evaluated on interpolated values. This test case indicates that FNODE can learn the frictional dynamics and approximate the resulting non-smooth acceleration. To improve fidelity near these non-smooth transitions, one can increase FNODE model complexity (e.g., deeper/wider hidden layers) to better represent piecewise-smooth behavior.
}

\begin{figure}[htbp]
	\centering
	\includegraphics[width=12cm]{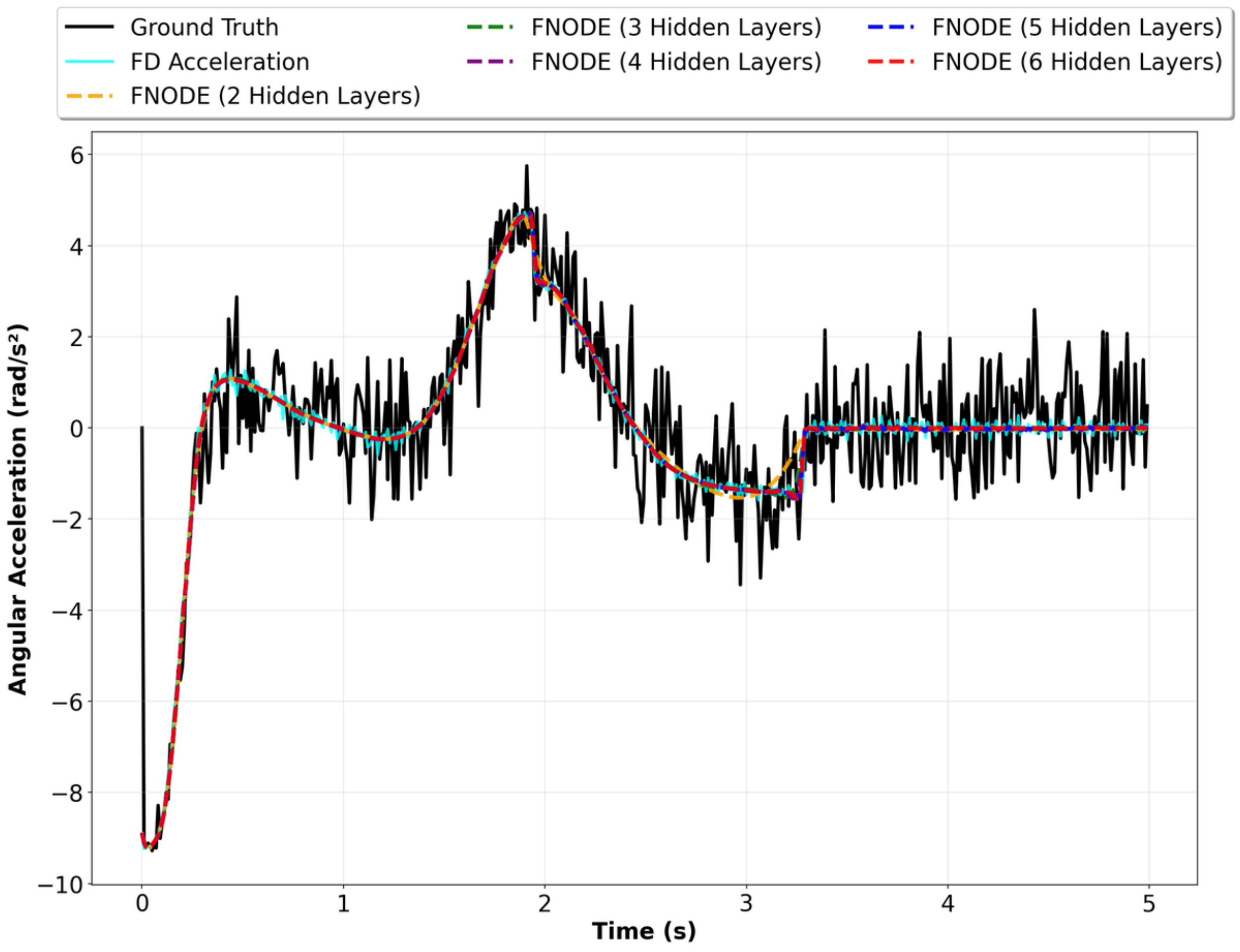}
	\caption{Acceleration prediction for the slider-crank mechanism ($\mu=0.6$) using FNODE models with varying depths.}
	\label{fig:sc_acc_approx}
\end{figure}

\updatedText{A central challenge for FNODE modeling in the presence of friction is accurately predicting the resulting non-smooth acceleration field. A common way to improve accuracy in such settings is to increase network depth, which enhances the model’s ability to represent sharp, piecewise-smooth transitions. Figure~\ref{fig:sc_acc_approx} shows FNODE acceleration approximations for the slider-crank mechanism at $\mu=0.6$ using models with different numbers of hidden layers. As depth increases, the approximation consistently improves, as evidenced by the decreasing MSE values reported in Table~\ref{tab:slider_crank_accel}.}

\begin{table}[htbp]
	\centering
	\caption{\updatedText{MSE of various-layers FNODE acceleration prediction for the slider-crank mechanism with friction coefficient $\mu=0.6$}}
	{
	\setlength{\tabcolsep}{6pt}
	\renewcommand{\arraystretch}{1.2}
	\begin{tabular}{@{}cc@{}}
		\toprule
		Number of Hidden Layers & MSE \\
		\midrule
		2 & 3.70e-02 \\
		3 & 1.72e-02 \\
		4 & 1.47e-02 \\
		5 & 1.44e-02 \\
		6 & 1.41e-02 \\
		\bottomrule
	\end{tabular}
	}
	\label{tab:slider_crank_accel}
\end{table}


\subsection{Cart-Pole System}

\begin{figure}[htbp]
    \centering
    \includegraphics[width=12cm]{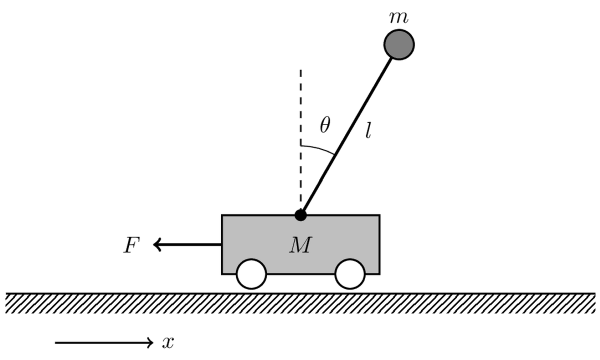}
    \caption{Cart-pole system.}
    \label{fig:cart_pole}
\end{figure}

\label{sec:cart-pole}
We conclude with the cart-pole system, considered here because it is a canonical control benchmark (Figure~\ref{fig:cart_pole}). The cart can move horizontally on a frictionless track, while the pendulum swings freely about its pivot, with motion restricted to $[-\pi/2,\pi/2]$. The system state is described by four variables: cart position $x$, velocity $\dot{x}$, pendulum angle $\theta$, and angular velocity $\dot{\theta}$. The dynamics are governed by nonlinear ODEs:
\begin{equation}
	\label{eqn_cart_pole}
	\begin{split}
		ml^2\ddot{\theta} + ml \cos \theta \ddot{x} - mgl\sin \theta & = 0 \\
		ml\ddot{\theta} \cos \theta  + (M + m)\ddot{x} - ml\dot{\theta}^2 \sin \theta & = u
	\end{split}
\end{equation}
The cart-pole was first tested in free evolution (no external input), initialized at
\[
x(0)=1,\; \dot{x}(0)=0,\; \theta(0)=\tfrac{\pi}{6},\; \dot{\theta}(0)=0.
\]
Trajectories were generated with a midpoint integrator at $\Delta t=0.01$ s. The first 200 steps (2 s) were used for training and the following 50 steps (0.5 s) for testing. Model hyperparameters are listed in Table~\ref{tab:hyper_cp}.

\begin{table}[h]
	\centering
	\caption{Hyper-parameters for the cart-pole system.}
	\label{tab:hyper_cp}
	\begin{tabular}{@{}lccccc@{}}
		\toprule
		Hyper-parameters & \multicolumn{4}{c}{ Model} \\
		\cmidrule(r){2-5}
		& FNODE & MBD-NODE          & LSTM &  FCNN\\ \midrule
		No. of hidden layers     &  2 & 2          & 2                          &2                \\
		No. of nodes per hidden layer & 256  & 256  & 256  & 256   \\
		Max. epochs  &  450      &  400       & 300        & 450               \\
		Initial learning rate & 1e-3 & 1e-3 &1e-3  & 1e-3 \\
		Learning rate decay &  0.98 &  0.98  &0.98    &  0.98       \\ 
		Activation function &  tanh &  tanh & tanh  &tanh \\
		Loss function     & MSE  & MSE &  MSE &  MSE                   \\
		Optimizer     &  Adam    &  Adam     & Adam & Adam                                   \\  \bottomrule
	\end{tabular}
\end{table}

\begin{figure}
	\centering
	\includegraphics[width=12cm]{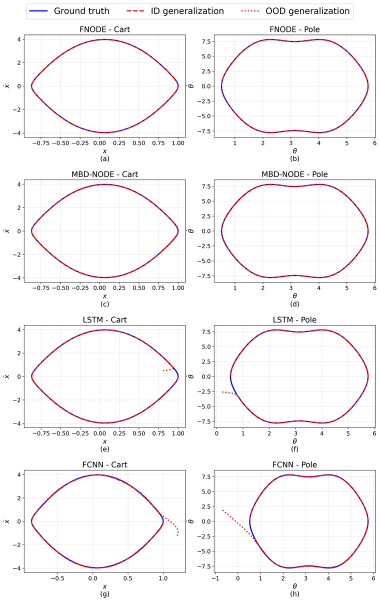}
	\caption{Phase-space trajectories for the cart-pole free-evolution scenario. The left subplot shows the cart's phase portrait ($\dot{x}$ vs. $x$), while the right subplot shows the pole's phase portrait ($\dot{\theta}$ vs. $\theta$). Dashed lines represent the fit to training data; dotted lines represent test data predictions. The models are: (a,b) FNODE, (c,d) MBD-NODE, (e,f) LSTM, and (g,h) FCNN.}
	\label{fig:cp_xv}
\end{figure}

\begin{figure}
	\centering
	\includegraphics[width=12cm]{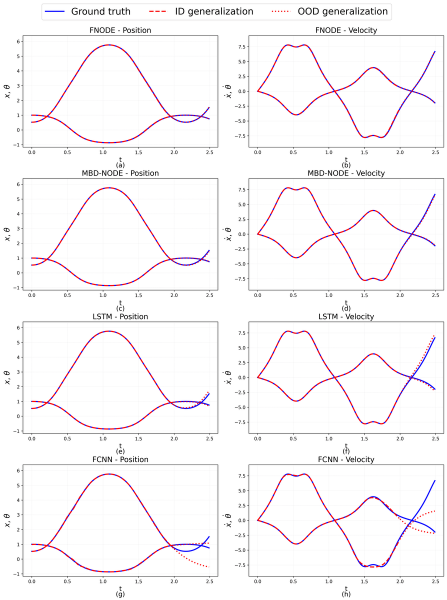}
	\caption{State variable trajectories over time for the cart-pole system. The left column shows position variables ($x, \theta$), and the right column shows velocity variables ($\dot{x}, \dot{\theta}$). Dashed lines correspond to the training interval ($t \in [0,2]$), and dotted lines to the test interval. The final MSEs are: FNODE ($\epsilon=6.7\text{e-}4$), MBD-NODE ($\epsilon=1.1\text{e-}3$), LSTM ($\epsilon=1.6\text{e-}1$), and FCNN ($\epsilon=3.5\text{e-}1$).}
	\label{fig:cp_xvt}
\end{figure}

Figures~\ref{fig:cp_xv}-\ref{fig:cp_xvt} compare model performance. FNODE and MBD-NODE both match the ground truth closely, with low MSEs ($6.7\!\times\!10^{-4}$ and $1.1\!\times\!10^{-3}$, respectively). MBD-NODE is slightly more accurate, but FNODE remains competitive while training far more efficiently. LSTM largely reproduces training patterns, which fit the periodic pattern of cart-pole system, so it still provides relatively accurate predictions. FCNN diverges significantly in extrapolation.

We also test FNODE in a model predictive control (MPC) task, where an external force $u$ is applied to stabilize the pole and return the cart to the origin. The MPC strategy, for control purposes, linearizes the dynamics and then solves a quadratic program over a finite horizon \cite{rawlings2020model}. For a linearized system $\dot{z}=Az+Bu$, the optimization problem is formulated as:
\begin{equation} \label{eq:mpc}
	\begin{aligned}
		\min_{u} \quad & \sum_{k=0}^{N-1} z_k^T Q z_k + u_k^T R u_k \\
		\text{s.t.} \quad & z_{k+1} = (I+\Delta t A) z_k + \Delta tB u_k, \quad k = 0,1,\dots,N-1 \\
		& z_k \in \mathcal{Z}, \quad u_k \in \mathcal{U}, \quad k = 0,1,\dots,N-1
	\end{aligned}
\end{equation}

At each step, the system state is $z_k=(\theta_k, x_k, \dot{\theta}_k, \dot{x}_k)^T$ with control input $u_k$. The MPC problem is solved over a horizon of length $N$, subject to state and input constraints $\mathcal{Z}$ and $\mathcal{U}$. The cost function weights are $Q$ and $R$, both chosen as identity matrices.
$A$ and $B$ are the linearized system matrices, obtained from the Jacobian of the dynamics. 

To solve the control problem, FNODE learns the mapping from state and control input to accelerations, $f(\theta_k, x_k, \dot{x}_k, \dot{\theta}_k, u) \mapsto (\ddot{\theta}_k, \ddot{x}_k)$. We use automatic differentiation and the inherent differentiable nature of a Neural Network to obtain the linearization matrices $A$ and $B$ required by the MPC framework. The standard (non-NN based) MPC for Cart-Pole is also provided as a baseline for comparison, see \ref{app:cart-pole-mpc} for the expression of the linearization matrices.

\begin{figure}
	\centering
	\includegraphics[width=12cm]{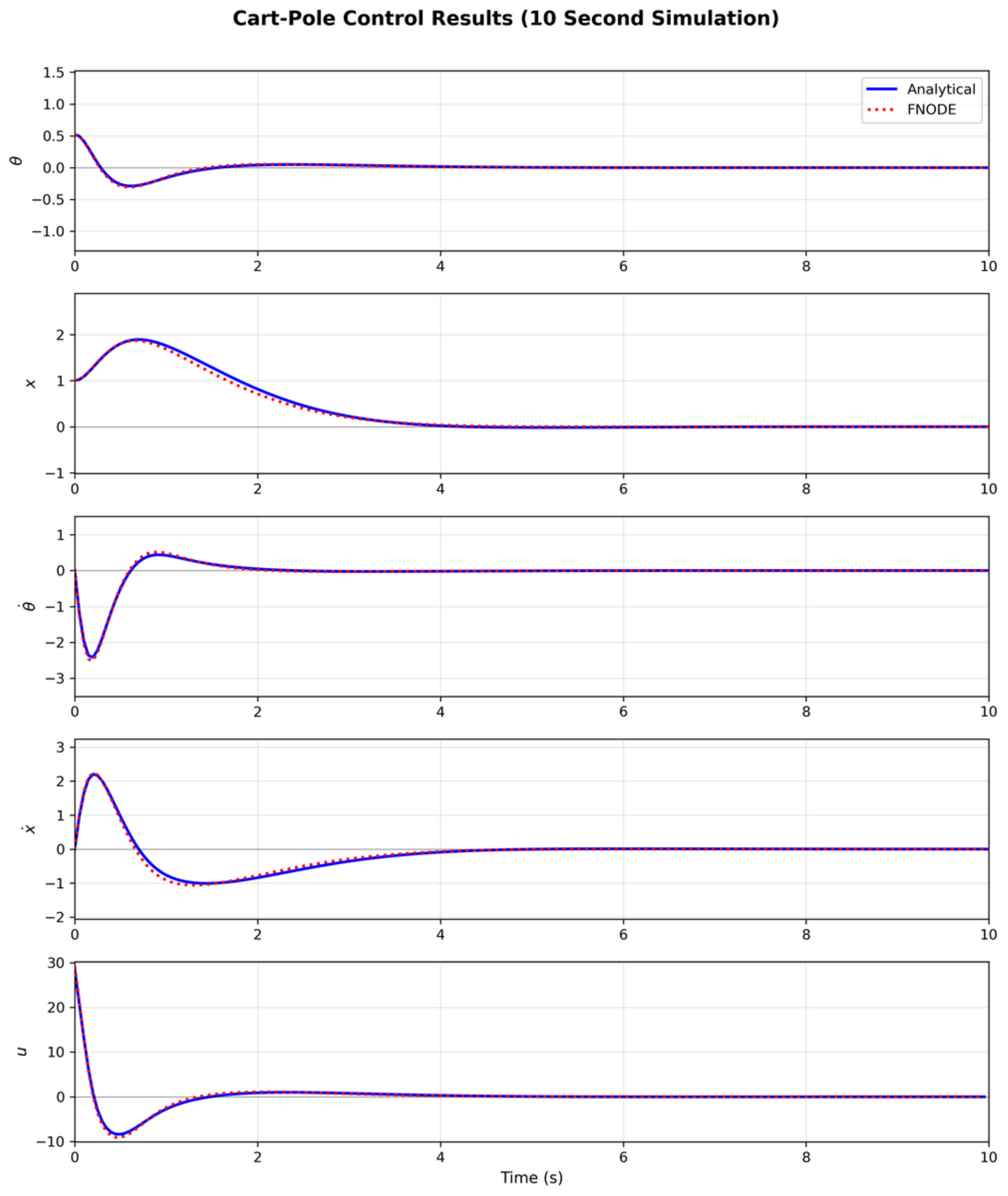}
	\caption{Comparative performance of the MPC strategy. The analytical model (solid blue) and FNODE-based model (dotted red) are compared across: (a) pole angle $\theta$, (b) cart position $x$, (c) pole angular velocity $\dot{\theta}$, (d) cart velocity $\dot{x}$, and (e) the control force $u$.}
	\label{fig:cp_mpc}
\end{figure}

Figure~\ref{fig:cp_mpc} shows the MPC results for the cart-pole. The five subplots depict the pole angle $\theta$, cart position $x$, angular velocity $\dot{\theta}$, cart velocity $\dot{x}$, and control input $u$. The FNODE-based controller (red dotted) closely matches the analytical baseline (blue solid), demonstrating that FNODE learns the dynamics with sufficient accuracy for supporting an MPC controller.

\updatedText{\subsection{Parameterized Vehicle Model}\label{sec:vehicle_4dof}}
\updatedText{We finally consider the simplified 4DOF vehicle model proposed in \cite{TR-2023-06,huzaifaMSthesis2023}. The model has a simple powertrain component and adopts throttle and steering as control inputs. The state and control vectors are:}

\updatedText{
\begin{equation}
	q = [x,\, y,\, \theta,\, v]^T,\qquad u=[\alpha,\,\delta]^T,
\end{equation}
where $(x,y)$ is the vehicle position in the global frame, $\theta$ is the heading (yaw) angle, $v$ is the longitudinal speed, $\alpha$ is the normalized throttle, and $\delta$ is the steering angle.
}
\updatedText{
The state transition model is
\begin{equation}\label{eq:veh4dof}
	\dot{q}=
	\begin{bmatrix}
		\dot{x}\\ \dot{y}\\ \dot{\theta}\\ \dot{v}
	\end{bmatrix}
	=
	\begin{bmatrix}
		\cos\theta \cdot v\\
		\sin\theta \cdot v\\
		\dfrac{v\tan\beta\delta}{l}\\[4pt]
		\dfrac{2\,T(\alpha,v)\,\gamma}{I_{\mathrm{wheel}}}\,R_{\mathrm{wheel}}
	\end{bmatrix},
\end{equation}
where $l$ is the wheelbase, $\beta$ maps steering command $\delta$ to a physical wheel steering angle, $\gamma$ is the gear ratio, and $I_{\mathrm{wheel}}$ and $R_{\mathrm{wheel}}$ are the wheel inertia and radius. The motor torque is modeled as
\begin{equation}\label{eq:veh4dof_torque}
	\begin{aligned}
		T(\alpha,v) &= T'(\alpha,\omega_m)=\alpha f_1(\omega_m)-c_1\omega_m-c_0 \;, \\
		f_1(\omega_m) &= -\frac{\tau_0}{\omega_0}\omega_m+\tau_0 \;, \\
		\omega_m &= \frac{v}{R_{\mathrm{wheel}}\gamma} \;,
	\end{aligned}
\end{equation}
where $\tau_0$ and $\omega_0$ are the stalling torque and no-load speed of the motor, and $c_0,c_1$ model the motor resistance torque.
}
\updatedText{
This example contains nine scalar parameters
\begin{equation}
\label{eq:veh4dof_params}
	\Theta_p = [l,\,\beta,\,\gamma,\,I_{\mathrm{wheel}},\,R_{\mathrm{wheel}},\,c_0,\,c_1,\,\tau_0,\,\omega_0]^T \;,
\end{equation}
each sampled on a grid with four values, yielding $4^9$ parameter combinations. We simulate each parameter instance for 20 s with time step $\Delta t=0.01$ (2000 steps). The control signal $u(t)=[\alpha(t),\delta(t)]^T$ is pre-defined and shared by all parameter instances; the corresponding state trajectories are generated by the RK4 integrator. For each trajectory, the first 1500 steps are used for training and the remaining 500 steps for testing (extrapolation).
}

\begin{table}[htbp]
	\centering
	\caption{\updatedText{Parameter ranges for the 4DOF vehicle model}}
	\label{tab:veh4dof_param_ranges}
	{
	\begin{tabular}{@{}lcc@{}}
		\toprule
		Parameter & Range & Explanation \\ 
		\midrule
		$l$ & \text{[2.7, 2.9]} & wheelbase \\
		$\beta$ & \text{[0.60, 0.75]} & steering ratio \\
		$\gamma$ & \text{[0.15, 0.25]} & gear ratio \\
		$I_{\mathrm{wheel}}$ & \text{[0.9, 1.3]} & wheel inertia \\
		$R_{\mathrm{wheel}}$ & \text{[0.32, 0.35]} & wheel radius \\
		$c_0$ & \text{[0.015, 0.025]} & motor resistance torque coefficient \\
		$c_1$ & \text{[0.025, 0.045]} & motor resistance torque coefficient \\
		$\tau_0$ & \text{[250, 350]} & stalling torque \\
		$\omega_0$ & \text{[1200, 1500]} & no-load speed \\
		\bottomrule
	\end{tabular}
	}
\end{table}

\begin{table}[htbp]
	\centering
	\caption{\updatedText{Test parameters for the four test cases.}}
	\label{tab:veh4dof_param_testcases}
	\setlength{\tabcolsep}{6pt}
	\renewcommand{\arraystretch}{1.2}
	{
	\begin{tabular}{@{}cccccccccc@{}}
		\toprule
		Test & $l$ & $R_{\mathrm{wheel}}$ & $I_{\mathrm{wheel}}$ & $\tau_0$ & $\omega_0$ & $c_0$ & $c_1$ & $\beta$ & $\gamma$ \\
		\midrule
		1 & 2.7 & 0.32 & 0.9 & 250 & 1200 & 0.015 & 0.025 & 0.6 & 0.15 \\
		2 & 2.9 & 0.35 & 1.3 & 350 & 1500 & 0.025 & 0.045 & 0.75 & 0.25 \\
		3 & 2.726 & 0.3281 & 0.952 & 277 & 1239 & 0.0177 & 0.0276 & 0.6405 & 0.163 \\
		4 & 2.846 & 0.3461 & 1.192 & 337 & 1419 & 0.0237 & 0.0396 & 0.7305 & 0.223 \\
		\bottomrule
	\end{tabular}
	}
\end{table}

\updatedText{FNODE learns the mapping from the full state and control inputs to the state time derivative, i.e., $(q,u,\Theta_p)\mapsto\dot{q}$ as defined in Eqs.~\eqref{eq:veh4dof}--\eqref{eq:veh4dof_params}. All models are trained on the same trajectories and evaluated on the held-out extrapolation interval. The hyperparameters are summarized in Table~\ref{tab:hyper_veh4dof}.
}

\begin{table}[htbp]
	\centering
	\caption{\updatedText{Hyper-parameters for the 4DOF vehicle model.}}
	\label{tab:hyper_veh4dof}
	{
	\begin{tabular}{@{}lccccc@{}}
		\toprule
		Hyper-parameters & \multicolumn{1}{c}{Model} \\
		\cmidrule(r){2-2}
		& FNODE \\ \midrule
		No. of hidden layers     & 6            \\
		No. of nodes per hidden layer &256 \\
		Max. epochs            &\textit{300}      \\
		Initial learning rate &1e-4 \\
		Learning rate decay &0.98 \\
		Activation function &tanh \\
		Loss function        &MSE   \\
		Optimizer              &AdamW   \\  \bottomrule
	\end{tabular}
	}
\end{table}

\begin{figure}
	\centering
	\includegraphics[width=12cm]{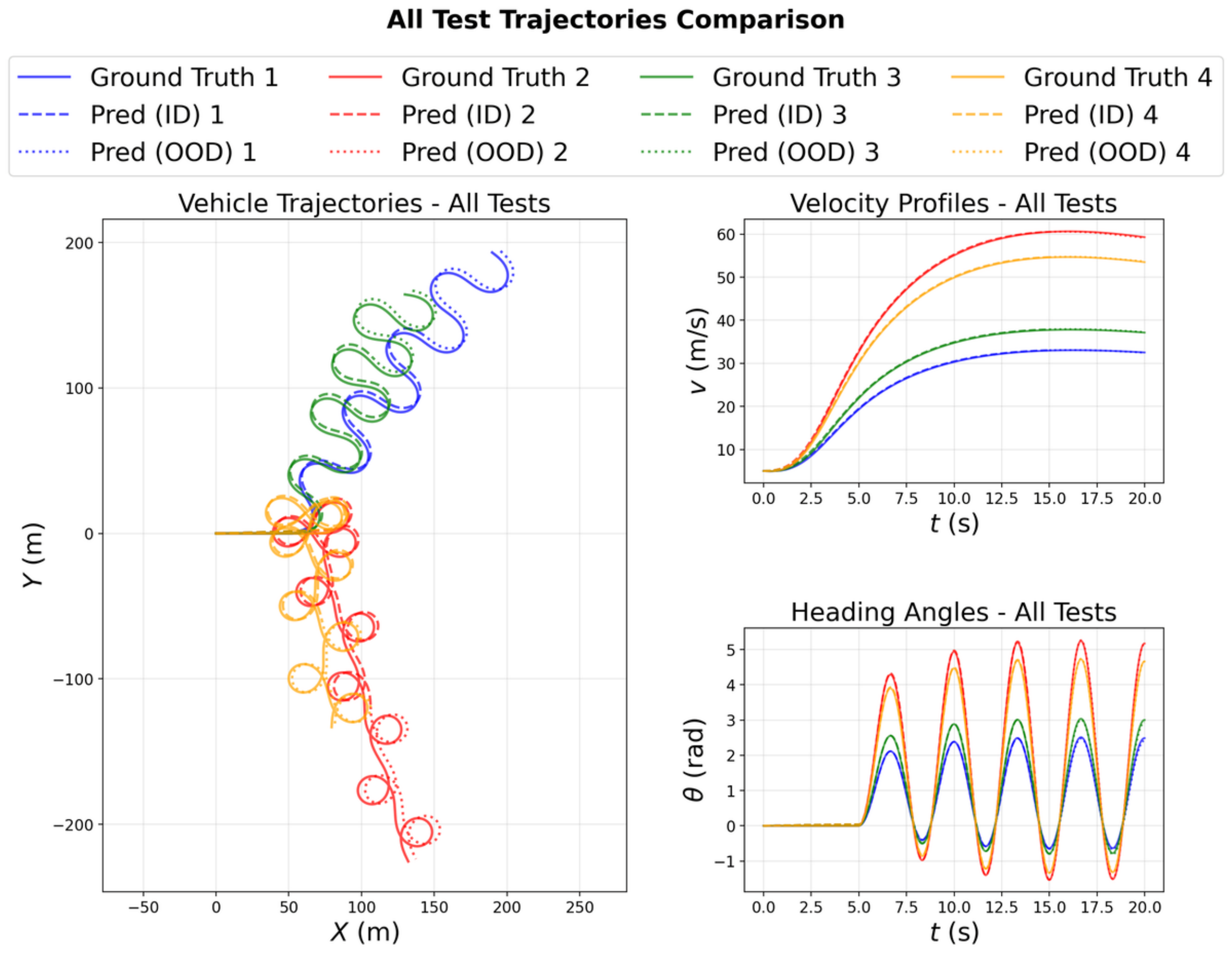}
	\caption{\updatedText{Trajectory Comparison of the 4DOF vehicle model.}}
	\label{fig:veh4dof}
\end{figure}

\updatedText{Figure~\ref{fig:veh4dof} compares the predicted trajectories of $q$ against the ground truth under identical control inputs. 
Tests 1-2 are drawn from the corners of the parameter grid, whereas tests 3-4 lie in the interior (Table~\ref{tab:veh4dof_param_testcases}) 
and therefore assess interpolation in parameter space. All methods track well within the training interval; discrepancies arise under extrapolation, 
where integration errors accumulate and parameter variations compound. FNODE remains stable across the grid and delivers comparatively accurate 
predictions over the test window, while the baselines exhibit larger drift and occasional divergence. Because the system's high dimensionality 
makes the baseline datasets prohibitively large, we report only FNODE results here. Future work will investigate scalable baseline 
implementations for high-dimensional parameter spaces.}

\FloatBarrier

\section{Discussion}
\label{sec:discussion}
\begin{figure}[h]
	\centering
   \includegraphics[width=12cm]{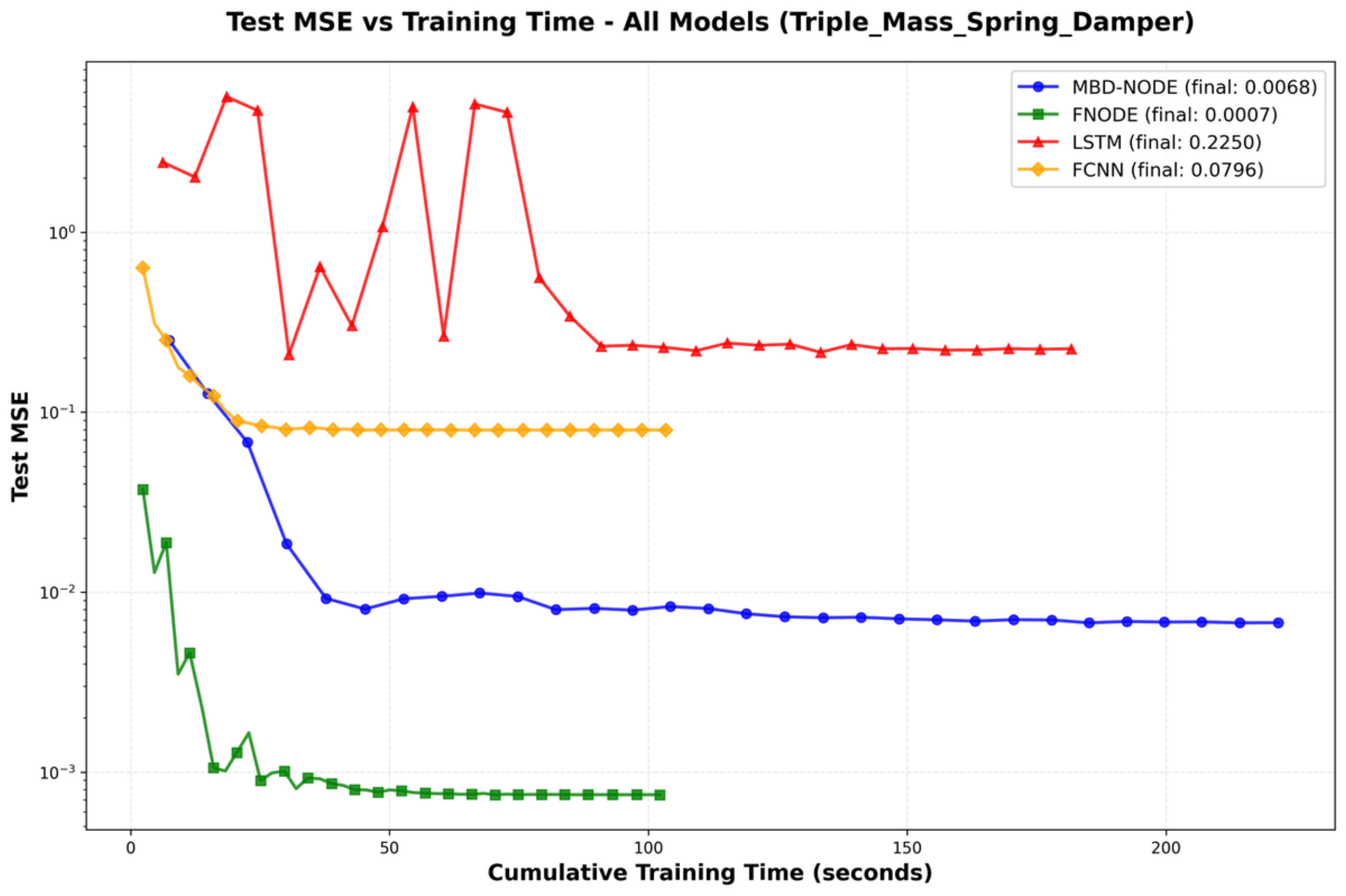}
	\caption{The test MSE vs training time comparison of the all models on the triple mass-spring-damper benchmark.}
	\label{fig:tmsd_mse_comparison}
\end{figure}

Figure~\ref{fig:tmsd_mse_comparison} summarizes the computational cost and predictive accuracy of the four models for the triple mass-spring-damper benchmark. Two observations emerge.

First, the training process of FNODE converges with lower computational cost than MBD-NODE.
The computational cost is quantified by the training time of the model. As shown in Figure~\ref{fig:tmsd_mse_comparison},
the train loss of FNODE converges with lower training time compared to MBD-NODE.
This gap arises because FNODE bypasses the adjoint sensitivity method and does not embed an
ODE solver in the backpropagation loop. Compared to purely data-driven
baselines such as FCNN, FNODE's training times are of the same order of magnitude.

Second, FNODE achieves consistently lower prediction error than the black-box baselines and is competitive with, 
or superior to, MBD-NODE. In dissipative and chaotic systems (single/triple mass-spring-damper, double pendulum), 
FNODE outperforms all baselines by large margins, maintaining stable long-term predictions. For the slider-crank mechanism, 
FNODE achieves the best accuracy ($1.8\!\times\!10^{-3}$ MSE), significantly outperforming both MBD-NODE and the 
data-driven models. Only in the cart-pole case does MBD-NODE obtain a slightly lower error than FNODE. 
These results underscore the importance of accurate acceleration 
targets: FFT provides precision for smooth trajectories, while FD ensures robustness for non-periodic signals 
and trajectory endpoints.

FNODE provides a good efficiency vs. accuracy balance: it is orders of magnitude faster 
to train than MBD-NODE while delivering consistently higher fidelity than LSTM and FCNN, and near-parity or 
better accuracy than MBD-NODE across benchmarks. \updatedText{These results suggest that FNODE is computationally practical for higher-dimensional constrained multibody systems and can accommodate parameter variation, as illustrated by the added higher-DOF vehicle example and parameter-conditioned experiments.}

\updatedText{In its current form, FNODE is best suited to systems with smooth or piecewise-smooth trajectories, where accelerations are well-defined and can be estimated reliably from sampled data (here via FFT/FD differentiation). Our results indicate that, despite the resulting non-smoothness, a Coulomb friction model remains amenable to FNODE, which can learn an accurate approximation of the associated acceleration field. In contrast, genuinely nonsmooth events, e.g., impacts (chattering) and complementarity-based stick-slip transitions, are deferred to future work, as discussed in the Conclusions.
}

\begin{table}[htbp]
 \centering
 \begin{tabular}{|m{4cm}|m{2cm}|m{2cm}|m{3.0cm}|}
    \hline
    \textbf{Test Case} & \textbf{Model}& 
    \textbf{Integrator}& \textbf{Time Cost (s)} \\ \hline

    \multirow{4}{4cm}{Single Mass Spring Damper} 
    & FNODE  & -- &  \updatedText{112.46} \\ \cline{2-4} 
    & MBD-NODE  & RK4 & \updatedText{201.90} \\ \cline{2-4} 
    & FCNN  & -- & \updatedText{103.28} \\ \cline{2-4}
    & LSTM  & -- & \updatedText{171.99} \\ \hline
    
    \multirow{4}{4cm}{Triple Mass Spring Damper} 
    & FNODE& -- & \updatedText{108.23} \\ \cline{2-4}
    & MBD-NODE  & RK4 & \updatedText{244.80} \\ \cline{2-4} 
    & FCNN  & -- & \updatedText{102.78} \\ \cline{2-4}
    & LSTM  & -- & \updatedText{172.05} \\ \hline
   
    \multirow{4}{4cm}{Double Pendulum} 
    & FNODE  & -- & 89.33 \\ \cline{2-4} 
    & MBD-NODE  & RK4 & 212.09 \\ \cline{2-4} 
    & FCNN  & -- & 88.04 \\ \cline{2-4} 
    & LSTM  & -- & 148.41 \\ \hline
    
    \multirow{4}{4cm}{Slider Crank}
    & FNODE  & -- & \updatedText{565.01 }\\ \cline{2-4} 
    & MBD-NODE  & RK4 & \updatedText{1185.64} \\ \cline{2-4} 
    & FCNN  & -- & \updatedText{453.50} \\ \cline{2-4} 
    & LSTM  & -- & \updatedText{813.11} \\ \hline

    \multirow{4}{4cm}{Cart Pole}
    & FNODE  & -- & 59.46 \\ \cline{2-4} 
    & MBD-NODE  & RK4 & 144.51 \\ \cline{2-4} 
    & FCNN  & -- & 59.36 \\ \cline{2-4} 
    & LSTM  & -- & 96.08 \\ \hline

    \multirow{4}{4cm}{4dof Vehicle}
    & FNODE  & -- & 3304.32 \\ \cline{2-4} 
    & MBD-NODE  & RK4 & -- \\ \cline{2-4} 
    & FCNN  & -- & -- \\ \cline{2-4} 
    & LSTM  & -- & -- \\ \hline
 \end{tabular}
 \label{tab:time_cost}
 \caption{Training time cost for the models. RK4 denotes the 4th-order Runge-Kutta method. The models are trained on Nvidia GeForce 5070Ti GPU with 16 GB memory.}
\end{table}

\begin{table}[h]
\centering
\begin{tabular}{lcccc}
\toprule
\textbf{Test Case} & \multicolumn{4}{c}{\textbf{Error}} \\
\cmidrule(lr){2-5}
 & \textbf{FNODE} & \textbf{MBD-NODE} & \textbf{LSTM} & \textbf{FCNN} \\
\midrule
\updatedText{Single Mass-Spring-Damper} & \updatedText{7.6e-5} & \updatedText{1.4e-3} & \updatedText{1.5e-2} & \updatedText{1.6e-1} \\
\updatedText{Triple Mass-Spring-Damper} & \updatedText{6.1e-3} & \updatedText{6.8e-3} & \updatedText{2.5e-1} & \updatedText{9.0e-2} \\
Double Pendulum & 1.4e-1 & 2.3e-1 & 7.6e-1 & 1.9e0 \\
\updatedText{Slider Crank} & \updatedText{1.7e-1} & \updatedText{8.7e-1} & \updatedText{1.4e0} & \updatedText{3.1e0} \\
Cart Pole & 6.7e-4 & 1.1e-3 & 1.6e-1 & 3.5e-1 \\
\bottomrule
\end{tabular}
\label{tab:error_summary}
\caption{Summary of the numerical MSE for different models. }
\end{table}

\FloatBarrier

\section{Conclusions and Future Work}
\label{sec:conclusions}
We introduced FNODE, a framework for data-driven modeling of multibody dynamics that learns accelerations directly 
rather than states. By reformulating the training objective and leveraging FFT- and finite-difference-based differentiation, 
FNODE eliminates the adjoint bottleneck of traditional Neural ODEs. Across benchmarks ranging from simple oscillators to 
chaotic and constrained mechanisms, FNODE trained one to two orders of magnitude faster than MBD-NODE while outperforming 
black-box baselines such as LSTM and FCNN. Key outcomes include:
\begin{itemize}
    \item Good accuracy and generalization: FNODE achieved lower MSE than all baselines, performing well in both interpolation and extrapolation.
    \item Data efficiency: Accurate long-term predictions were obtained from short training trajectories, e.g., thousands of steps extrapolated in the slider-crank test.
    \item Multiscale robustness: FNODE captured dynamics across time scales, as shown in the triple-mass-spring-damper system.
    \item Reproducibility: Our open-source implementation and benchmark suite, which includes all NNs and dynamics models used in this study, are provided for reproducibility studies and further research.
\end{itemize}

\updatedText{FFT-based spectral differentiation yields high accuracy for smooth, uniformly sampled trajectories when periodicity (or approximate periodicity) is a reasonable assumption, since the derivative is obtained from a global Fourier representation. However, for finite, non-periodic trajectories the implicit periodic extension can introduce endpoint artifacts (spectral leakage/Gibbs-type effects), motivating the use of local finite-difference (FD) stencils near boundaries. The FD schemes employed in this work were of lower order (e.g., $\mathcal{O}(\Delta t^2)$ in the interior and $\mathcal{O}(\Delta t)$ at endpoints for the stencils used here), but are robust for truncated data and do not rely on periodicity. Higher order FD, which have not been tested in this work, could be more accurate but may amplify noise. By employing the FFT and FD schemes in a complementary way, i.e., FFT where its assumptions are approximately satisfied and FD where boundary effects or non-periodicity dominate, we balance accuracy and robustness without relying on either method exclusively.}

\updatedSecond{In the MSE comparison test in Section~\ref{sec:single_mass_spring}, we compared the MSE of Neural Network learned field mismatch and the error of numerical differentiation in the same setting. 
The result shows that the error of Neural Network learned field mismatch is dominated by the error of numerical differentiation. 
This suggests that with accurate acceleration calculated by numerical differentiation, FNODE is able to achieve good accuracy without numerical integrator in training process.}

\updatedText{There are several limitations of the approach which we stopped short of addressing and thus are deferred to future work. First, FNODE relies on accurate acceleration estimates from trajectory data; both FFT and FD methods can suffer from noise and, in the case of FFT, Gibbs artifacts. Further work should be carried out to investigate the sensitivity of the approach to the choice of differentiation scheme. Second, integration error compounds over time, reducing long-term accuracy in chaotic systems such as the double pendulum. However, even when this is the case, the proposed approach still yields accurate predictions for short-term and intermediate-term horizons, a key requirement for controls applications. }

\updatedText{Future work should investigate also noise-resilient 
methods for estimating accelerations from sparse or imperfect data. The goal is to reduce long-term error growth, for example by embedding physical constraints or conservation laws into training or integration, in line with the methodology embraced for the slider-crank example, which has kinematic constraints. 
An important direction for future work is a systematic numerical study of differentiation-induced errors, including evaluation of the stencil-disagreement indicator $\eta'$ and the trajectory reconstruction residual $\eta_z$ across datasets and model variants. Finally, one rich direction of research is extending FNODE to problems with friction and contact, in the vein of the slider-crank example. Handling friction is expected to be relatively straightforward for smoothed, penalty-based formulations in which friction is represented through history-dependent `bristle' dynamics, as in models such as LuGre and Dahl. The problem is more challenging for complementarity-based or Lagrange-multiplier formulations, where stick-slip transitions are enforced non-smoothly and can introduce abrupt regime changes. Beyond multibody dynamics, several avenues for future work include applying FNODE ideas to PDE-governed systems, integrating uncertainty quantification, and exploring closed-loop control applications.}

\updatedText{The code and scripts used for this work are publicly available as open source for reproducibility studies and further research \cite{FNODE_supportData2025}.}

\section*{Acknowledgements}
This work was carried out in part with support from National Science Foundation project CMMI2153855.

\section*{Declaration of Generative AI and AI-assisted technologies in the writing process}
During the preparation of this work the authors used OpenAI ChatGPT in order to improve language and readability of the text. After using this service, the authors reviewed and edited the content as needed and take full responsibility for the content of the publication.

\section*{Conflict of interest}
The authors declare that they have no known competing financial interests or personal relationships that could have appeared to influence the work reported in this paper.

\bibliographystyle{plain}
\bibliography{compiled_refs} 
\newpage

\appendix

\section{Algorithms for training the FNODE}
\label{sec:alg}

\begin{algorithm}[htbp]
	\caption{Training Algorithm for FNODE without Constraints}
	\label{alg:fnode_no_constraints}
	\begin{algorithmic}[1]
		\State \textbf{Initialize:} FNODE network $f(\cdot;\boldsymbol{\Theta})$ predicting accelerations
		\State \textbf{Input:} Ground truth state--acceleration pairs $\mathcal{D} = \{(\mathbf{Z}_{i}, \ddot{\mathbf{z}}_i)\}_{i=0}^{T-1}$ where $\mathbf{Z}_i \in \mathbb{R}^{2n_b}$ and $\ddot{\mathbf{z}}_i \in \mathbb{R}^{n_b}$, optimizer and its settings
		\For{each epoch $e = 1, 2, \ldots, E$}
		\For{each training sample $(\mathbf{Z}_i, \ddot{\mathbf{z}}_i) \in \mathcal{D}$}
		\State Prepare input state $\mathbf{Z}_i = [\mathbf{z}_i, \dot{\mathbf{z}}_i]$ and target acceleration $\ddot{\mathbf{z}}_i$
		\State \textbf{Predict Acceleration:} $\hat{\ddot{\mathbf{z}}}_i = f(\mathbf{Z}_i;\boldsymbol{\Theta})$
		\State Compute loss $L = \|\ddot{\mathbf{z}}_i - \hat{\ddot{\mathbf{z}}}_i\|_2^2$
		\State Backpropagate the loss to compute gradients $\nabla_{\boldsymbol{\Theta}}L$
		\State Update the parameters using optimizer: $\boldsymbol{\Theta} = \text{Optimizer}(\boldsymbol{\Theta},\nabla_{\boldsymbol{\Theta}} L)$
		\EndFor
		\State Decay the learning rate using exponential schedule
		\EndFor
		\State \textbf{Output:} Trained FNODE $f(\cdot;\boldsymbol{\Theta}^*)$
	\end{algorithmic}
\end{algorithm}

\begin{algorithm}[htbp]
	\caption{Training Algorithm for FNODE with Constraints using Only Minimal Coordinates}
	\label{alg:fnode_constraints_minimal}
	\begin{algorithmic}[1]
		\State \textbf{Initialize:} FNODE network $f(\cdot;\boldsymbol{\Theta})$ for minimal coordinates predicting accelerations; Choose integrator $\Phi$, use prior knowledge of constraint equation $\varphi$ and the minimal coordinates.
		\State \textbf{Input:} Minimal coordinate state--acceleration pairs $\mathcal{D}^M = \{(\mathbf{Z}_i^M, \ddot{\mathbf{z}}_i^M)\}_{i=0}^{T-1}$ where $\mathbf{Z}_i^M \in \mathbb{R}^{2m}$ and $\ddot{\mathbf{z}}_i^M \in \mathbb{R}^{m}$, constraint equations $\varphi$, optimizer and its settings
		\For{each epoch $e = 1, 2, \ldots, E$}
		\For{each training sample $(\mathbf{Z}_i^M, \ddot{\mathbf{z}}_i^M) \in \mathcal{D}^M$}
		\State Prepare minimal state $\mathbf{Z}_i^M = [\mathbf{z}_i^M, \dot{\mathbf{z}}_i^M]$ and target acceleration $\ddot{\mathbf{z}}_i^M$
		\State \textbf{Predict Minimal Acceleration:} $\hat{\ddot{\mathbf{z}}}_i^M = f(\mathbf{Z}_i^M;\boldsymbol{\Theta})$
		\State Compute minimal loss $L = \|\ddot{\mathbf{z}}_i^M - \hat{\ddot{\mathbf{z}}}_i^M\|_2^2$ 
		\State Backpropagate the loss to compute gradients $\nabla_{\boldsymbol{\Theta}}L$
		\State Update the parameters using optimizer: $\boldsymbol{\Theta} = \text{Optimizer}(\boldsymbol{\Theta},\nabla_{\boldsymbol{\Theta}} L)$
		\EndFor
		\State Decay the learning rate using exponential schedule
		\EndFor
		\State \textbf{Output:} Trained FNODE $f(\cdot;\boldsymbol{\Theta}^*)$ for minimal coordinates
	\end{algorithmic}
\end{algorithm}

\section{Double Pendulum Hamiltonian Formulation}
\label{app:double_pendulum}

For completeness, we outline the Hamiltonian derivation of the double pendulum dynamics used in Section~\ref{sec:numericalexperiments}.	
	The Hamiltonian is obtained via a Legendre transform of the Lagrangian and is written as
	\begin{equation}
		H(\theta_1,\theta_2,p_{\theta_1},p_{\theta_2}) = T(\theta_1,\theta_2,\dot{\theta}_1,\dot{\theta}_2) + V(\theta_1,\theta_2),
	\end{equation}
	with kinetic and potential energy
	\begin{align}
		T &= \tfrac{1}{2} m_1 l_1^2 \dot{\theta}_1^2 + \tfrac{1}{2} m_2 \big(l_1^2 \dot{\theta}_1^2 + l_2^2 \dot{\theta}_2^2 + 2 l_1 l_2 \dot{\theta}_1 \dot{\theta}_2 \cos(\theta_1-\theta_2)\big), \\
		V &= -m_1 g l_1 \cos\theta_1 - m_2 g \big(l_1 \cos\theta_1 + l_2 \cos\theta_2\big).
	\end{align}
	The canonical momenta are defined by $p_{\theta_i}=\tfrac{\partial T}{\partial \dot\theta_i}$. Using the expression for $T$ above yields
	\begin{align}
		p_{\theta_1} &= (m_1+m_2)l_1^2\dot\theta_1 + m_2 l_1 l_2 \dot\theta_2 \cos(\theta_1-\theta_2), \\
		p_{\theta_2} &= m_2 l_2^2\dot\theta_2 + m_2 l_1 l_2 \dot\theta_1 \cos(\theta_1-\theta_2).
	\end{align}
	Inverting these relations eliminates $\dot\theta_1$ and $\dot\theta_2$ in favor of $p_{\theta_1}$ and $p_{\theta_2}$, producing $H(\theta_1,\theta_2,p_{\theta_1},p_{\theta_2})$ and the explicit Hamiltonian equations below.
	
	From Hamilton's equations
	\[
	\dot{\theta}_i = \frac{\partial H}{\partial p_{\theta_i}}, \qquad
	\dot{p}_{\theta_i} = -\frac{\partial H}{\partial \theta_i},
	\]
	we obtain the explicit form of the governing equations:
	\begin{align}
		\dot{\theta}_1 &= \frac{l_2 p_{\theta_1} - l_1 p_{\theta_2}\cos(\theta_1-\theta_2)}{l_1^2 l_2\big[m_1+m_2\sin^2(\theta_1-\theta_2)\big]}, \\
		\dot{\theta}_2 &= \frac{-m_2 l_2 p_{\theta_1}\cos(\theta_1-\theta_2) + (m_1+m_2) l_1 p_{\theta_2}}{m_2 l_1 l_2^2 \big[m_1+m_2\sin^2(\theta_1-\theta_2)\big]}, \\
		\dot{p}_{\theta_1} &= -(m_1+m_2) g l_1 \sin\theta_1 - h_1 + h_2 \sin\big(2(\theta_1-\theta_2)\big), \\
		\dot{p}_{\theta_2} &= -m_2 g l_2 \sin\theta_2 + h_1 - h_2 \sin\big(2(\theta_1-\theta_2)\big),
	\end{align}
	with auxiliary terms
	\begin{align}
		h_2 &= \frac{m_2 l_2^2 p_{\theta_1}^2 + (m_1+m_2) l_1^2 p_{\theta_2}^2 - 2 m_2 l_1 l_2 p_{\theta_1} p_{\theta_2} \cos(\theta_1 - \theta_2)}{2 l_1^2 l_2^2 \big[m_1+m_2 \sin^2(\theta_1-\theta_2)\big]^2}, \\
		h_1 &= \frac{p_{\theta_1} p_{\theta_2} \sin(\theta_1-\theta_2)}{l_1 l_2 \big[m_1+m_2 \sin^2(\theta_1-\theta_2)\big]}.
	\end{align}

\section{A slider-crank mechanism, with rigid bodies}
\label{sec:slider-crank-mechanism}

Based on the setup illustrated in Section \ref{sec:slider_crank}, we formulate the equation of motion of slider-crank mechanism:

The mass matrix $M \in \mathbb{R}^{9 \times 9}$ is:
\begin{equation}
		M = \begin{bmatrix}
			M_1 & 0_{3\times3} & 0_{3\times3} \\
			0_{3\times3} & M_2 & 0_{3\times3} \\
			0_{3\times3} & 0_{3\times3} & M_3
		\end{bmatrix},
\end{equation}
where:
	\begin{equation*}
		M_1 = \begin{bmatrix}
			m_1 & 0 & 0 \\
			0 & m_1 & 0 \\
			0 & 0 & I_1
		\end{bmatrix} = \begin{bmatrix}
			1 & 0 & 0 \\
			0 & 1 & 0 \\
			0 & 0 & 0.1
		\end{bmatrix},
	\end{equation*}

	\begin{equation*}
		M_2 = \begin{bmatrix}
			m_2 & 0 & 0 \\
			0 & m_2 & 0 \\
			0 & 0 & I_2
		\end{bmatrix} = \begin{bmatrix}
			1 & 0 & 0 \\
			0 & 1 & 0 \\
			0 & 0 & 0.1
		\end{bmatrix},
	\end{equation*}

	\begin{equation*}
		M_3 = \begin{bmatrix}
			m_3 & 0 & 0 \\
			0 & m_3 & 0 \\
			0 & 0 & I_3
		\end{bmatrix} = \begin{bmatrix}
			1 & 0 & 0 \\
			0 & 1 & 0 \\
			0 & 0 & 0.1
		\end{bmatrix}.
\end{equation*}
The states of the slider crank mechanism $(x_1,y_1,\theta_1,x_2,y_2,\theta_2,x_3,y_3,\theta_3)$ \updatedText{follow} the below constraints  $\Phi : \mathbb{R}^9 \rightarrow \mathbb{R}^8$ on the position:
\begin{equation}
	\label{constraint}
	\Phi(q) = \begin{bmatrix}
		x_1 - r\cos(\theta_1) \\
		y_1 - r\sin(\theta_1) \\
		x_1 + r\cos(\theta_1) - x_2 + l\cos(\theta_2) \\
		y_1 + r\sin(\theta_1) - y_2 + l\sin(\theta_2) \\
		x_2 + l\cos(\theta_2) - x_3 \\
		y_2 + l\sin(\theta_2) - y_3 \\
		y_3 \\
		\theta_3
	\end{bmatrix}.
\end{equation}

Then the constraints Jacobian assumes the following form:
\begin{equation}
	\Phi_q = \begin{bmatrix}
		1 & 0 & r\sin(\theta_1) & 0 & 0 & 0 & 0 & 0 & 0 \\
		0 & 1 & -r\cos(\theta_1) & 0 & 0 & 0 & 0 & 0 & 0 \\
		1 & 0 & -r\sin(\theta_1) & -1 & 0 & -l\sin(\theta_2) & 0 & 0 & 0 \\
		0 & 1 & r\cos(\theta_1) & 0 & -1 & l\cos(\theta_2) & 0 & 0 & 0 \\
		0 & 0 & 0 & 1 & 0 & -l\sin(\theta_2) & -1 & 0 & 0 \\
		0 & 0 & 0 & 0 & 1 & l\cos(\theta_2) & 0 & -1 & 0 \\
		0 & 0 & 0 & 0 & 0 & 0 & 0 & 1 & 0 \\
		0 & 0 & 0 & 0 & 0 & 0 & 0 & 0 & 1 \\
	\end{bmatrix}.
\end{equation}

The external forces vector $F_e \in \mathbb{R}^{9}$ is formulated as:
\begin{equation}
	F_e = \begin{bmatrix}
		F_{e1}\\
		F_{e2}\\
		F_{e3}
	\end{bmatrix},
\end{equation}

where:
\begin{subequations}
\updatedText{\begin{equation}
		F_{e1} = \begin{bmatrix}
			0 \\
			0 \\
			\tau-c_{01}\dot{\theta}_1
		\end{bmatrix} \in \mathbb{R}^{3},
\end{equation}
	\begin{equation}
		F_{e2} = \begin{bmatrix}
			0 \\
			-c_{12}(\dot{\theta}_1-\dot{\theta}_2)\\
			0 
		\end{bmatrix} \in \mathbb{R}^{3},
\end{equation}
\begin{equation}
	\label{eqn:F_e3_with_friction}
		F_{e3} = \begin{bmatrix}
			-k\Delta x_3-f-\mu |\lambda_7| -c\dot{x}_3\\
			0 \\
			0
		\end{bmatrix} \in \mathbb{R}^{3} \;,
\end{equation}
}
\end{subequations}
where $|\lambda_7|$ is the absolute value of the scalar Lagrange multiplier (reaction force) associated with the constraint $y_3 = 0$ in Eq.~(\ref{constraint}), and $\mu$ is the friction coefficient. 

The rearranged constraint equations on the acceleration can be formulated from (\ref{constraint}):
	\begin{subequations}
		\begin{align}
			\ddot{x}_1 + r\ddot{\theta}_1 \sin(\theta_1) + r\dot{\theta}_1^2 \cos(\theta_1) &= 0 \\
			\ddot{y}_1 - r\ddot{\theta}_1 \cos(\theta_1) + r\dot{\theta}_1^2 \sin(\theta_1) &= 0 \\
			\ddot{x}_1 - r\ddot{\theta}_1 \sin(\theta_1) - r\dot{\theta}_1^2 \cos(\theta_1) - \ddot{x}_2 - 2l \ddot{\theta}_2 \sin(\theta_2) - 2l \dot{\theta}_2^2 \cos(\theta_2) &= 0 \\
			\ddot{y}_1 + r\ddot{\theta}_1 \cos(\theta_1) - r\dot{\theta}_1^2 \sin(\theta_1) - \ddot{y}_2 + 2l \ddot{\theta}_2 \cos(\theta_2) - 2l \dot{\theta}_2^2 \sin(\theta_2) &= 0 \\
			\ddot{x}_2 - 2l \ddot{\theta}_2 \sin(\theta_2) - 2l \dot{\theta}_2^2 \cos(\theta_2) - \ddot{x}_3 &= 0 \\
			\ddot{y}_2 + 2l \ddot{\theta}_2 \cos(\theta_2) - 2l \dot{\theta}_2^2 \sin(\theta_2) - \ddot{y}_3 &= 0 \\
			\dot{\theta}_3 &= 0 \\
			\ddot{\theta}_3 &= 0
		\end{align}
	\end{subequations}

Then we can get the $\gamma_c$ as follows:
	\begin{equation}
		\gamma_c =\begin{bmatrix}
			-r\dot{\theta}_1^2 \cos(\theta_1) \\
			-r\dot{\theta}_1^2 \sin(\theta_1) \\
			r\dot{\theta}_1^2 \cos(\theta_1) + 2l\dot{\theta}_2^2 \cos(\theta_2) \\
			r\dot{\theta}_1^2 \sin(\theta_1) + 2l\dot{\theta}_2^2 \sin(\theta_2) \\
			2l\dot{\theta}_2^2 \cos(\theta_2) \\
			2l\dot{\theta}_2^2 \sin(\theta_2) \\
			0 \\
			0
		\end{bmatrix}.
\end{equation}

\section{Analytical Linearization for Cart-Pole}
\label{app:cart-pole-mpc}
Based on the setup in Section \ref{sec:cart-pole}, \updatedText{we formulate} the cart-pole as a nonlinear continuous system:
\begin{equation}
	\dot{z} = f(z, u)
\end{equation}

where $z = [\theta, x, \dot{\theta}, \dot{x}]^T$ is the system state, $u$ is the control input, $\dot{z} = [\dot{\theta}, \dot{x}, \ddot{\theta}, \ddot{x}]^T$ is the time derivative of $z$, and $f$ is the nonlinear function that defines the system dynamics.

To implement the MPC, we linearize the nonlinear system around the equilibrium point $(z_0, u_0)$ by first-order Taylor expansion:
\begin{equation}
\label{eqn:cart_pole_taylor}
	f(z, u) \approx f(z_0, u_0) + \frac{\partial f}{\partial z}\bigg|_{(z_0, u_0)}\Delta z + \frac{\partial f}{\partial u}\bigg|_{(z_0, u_0)}\Delta u
\end{equation}

In equilibrium state, $z_0 = 0, u_0 = 0, f(z_0, u_0) = 0$, and the deviations from equilibrium are $\Delta z = z - z_0 = z$ and $\Delta u = u - u_0 = u$. The linearized system in Eq.~\eqref{eqn:cart_pole_taylor} can be expressed as:

\begin{equation}
	\dot{z} = \frac{\partial f}{\partial z}\bigg|_{(z_0, u_0)} z + \frac{\partial f}{\partial u}\bigg|_{(z_0, u_0)} u
\end{equation}

Then we define the linearized system matrices $A$ and $B$, where

\begin{equation}
	\begin{aligned}
		A &= \frac{\partial f}{\partial z}\bigg|_{(z_0, u_0)}
		   = \begin{bmatrix}
			\frac{\partial \dot{\theta}}{\partial \theta} & \frac{\partial \dot{\theta}}{\partial x} & \frac{\partial \dot{\theta}}{\partial \dot{\theta}} & \frac{\partial \dot{\theta}}{\partial \dot{x}} \\[8pt]
			\frac{\partial \dot{x}}{\partial \theta} & \frac{\partial \dot{x}}{\partial x} & \frac{\partial \dot{x}}{\partial \dot{\theta}} & \frac{\partial \dot{x}}{\partial \dot{x}} \\[8pt]
			\frac{\partial \ddot{\theta}}{\partial \theta} & \frac{\partial \ddot{\theta}}{\partial x} & \frac{\partial \ddot{\theta}}{\partial \dot{\theta}} & \frac{\partial \ddot{\theta}}{\partial \dot{x}} \\[8pt]
			\frac{\partial \ddot{x}}{\partial \theta} & \frac{\partial \ddot{x}}{\partial x} & \frac{\partial \ddot{x}}{\partial \dot{\theta}} & \frac{\partial \ddot{x}}{\partial \dot{x}}
			\end{bmatrix} \\
		B &= \frac{\partial f}{\partial u}\bigg|_{(z_0, u_0)}
		   = \begin{bmatrix}
			\frac{\partial \dot{\theta}}{\partial u} \\[8pt]
			\frac{\partial \dot{x}}{\partial u} \\[8pt]
			\frac{\partial \ddot{\theta}}{\partial u} \\[8pt]
			\frac{\partial \ddot{x}}{\partial u}
			\end{bmatrix}
	\end{aligned}
\end{equation}

As it can be assumed that the pole remains near $\theta = 0$, we can linearize the system around this equilibrium point to obtain a linearized model for MPC design.
\begin{equation}
\label{eqn:cp_approx}
	\begin{aligned}
		\sin\theta &\approx \theta \\
		\cos\theta &\approx 1 \\
		\dot{\theta}^2 &\approx 0
	\end{aligned}
\end{equation}

Then substitute Eq.~\eqref{eqn:cp_approx} into the nonlinear system dynamics Eq.~\eqref{eqn_cart_pole} to obtain the linearized model.

\begin{equation}
	\label{eqn_cart_pole_linearized}
	\begin{split}
		ml^2\ddot{\theta} + ml \ddot{x} -mgl\theta& = 0 \\
		ml\ddot{\theta} + (M + m)\ddot{x} & = u
	\end{split}
\end{equation}

By rearranging Eq.~\eqref{eqn_cart_pole_linearized}, the accelerations $\ddot{\theta}$ and $\ddot{x}$ can be expressed as:

\updatedText{\begin{equation}
	\label{eqn:cart_pole_accel_linearized}
	\begin{aligned}
		\ddot{\theta} &= \frac{(M+m)g}{M l}\theta - \frac{1}{M l}u \\
		\ddot{x} &= -\frac{m g}{M}\theta + \frac{1}{M}u
	\end{aligned}
\end{equation}}

Thus, the linearized system matrices $A$ and $B$ can be derived as:

\updatedText{\begin{equation}
	A = \begin{bmatrix}
		0 & 0 & 1 & 0 \\
		0 & 0 & 0 & 1 \\
		\frac{(M+m)g}{M l} & 0 & 0 & 0 \\
		-\frac{m g}{M} & 0 & 0 & 0
	\end{bmatrix}, \quad
	B = \begin{bmatrix}
		0 \\
		0 \\
		-\frac{1}{M l} \\
		\frac{1}{M}
	\end{bmatrix}
\end{equation}}

\clearpage
\newpage

\end{document}